\documentclass[10pt,journal]{IEEEtran}
\usepackage{cite}
\usepackage{url}
\usepackage{ragged2e}
\usepackage{footnote}
\makesavenoteenv{tabular}
\makesavenoteenv{table}
\usepackage{epsfig}
\usepackage{graphicx}
\usepackage{amsmath,amssymb} 
\usepackage{algorithm}
\usepackage{algorithmicx}
\usepackage{algpseudocode}
\usepackage{graphics}
\usepackage{threeparttable}
\usepackage{color}
\usepackage[normalem]{ulem}
\usepackage{multirow}
\usepackage{float}
\usepackage{amsfonts}
\usepackage{bm}
\usepackage{array}
\usepackage{pifont}
\usepackage{diagbox}
\usepackage{rotating}
\usepackage{booktabs}
\usepackage{overpic}
\usepackage{textcomp}
\usepackage{contour}
\usepackage{makecell}

\usepackage{enumitem}
\usepackage{colortbl}
\usepackage[american]{babel}
\usepackage{microtype}
\usepackage{bbding}

\usepackage[pagebackref=false,breaklinks=true,colorlinks, bookmarks=false]{hyperref}

\graphicspath{{./Imgs/}{./Imgs/authors/}}
\DeclareGraphicsExtensions{.pdf,.jpg,.png}

\ifdefined \GramaCheck
  \newcommand{\CheckRmv}[1]{}
  \newcommand{\figref}[1]{Figure 1}%
  \newcommand{\tabref}[1]{Table 1}%
  \newcommand{\secref}[1]{Section 1}
  \newcommand{\equref}[1]{Equation 1}
  \newcommand{\algref}[1]{Algorithm 1}
\else
 \newcommand{\CheckRmv}[1]{#1}
  \newcommand{\figref}[1]{Fig.~\ref{#1}}
  \newcommand{\tabref}[1]{Table~\ref{#1}}
  \newcommand{\secref}[1]{Section~\ref{#1}}
  \newcommand{\equref}[1]{Equ. (\ref{#1})}
\fi

\usepackage{colortbl}
\usepackage{cleveref}
\crefformat{section}{\S#2#1#3}
\crefformat{subsection}{\S#2#1#3}
\crefformat{subsubsection}{\S#2#1#3}

\usepackage{silence}
\def\ie{\emph{i.e.}}
\def\eg{\emph{e.g.}}
\def\etc{\emph{etc}}
\def\etal{{\em et al.~}}
\def\sArt{{state-of-the-art~}}

\newcommand{\revise}[1]{{\textcolor{black}{#1}}}

\begin{document}

\title{Regularized Densely-connected Pyramid Network for Salient Instance Segmentation}

\author{Yu-Huan Wu, Yun Liu, Le Zhang, Wang Gao, and Ming-Ming Cheng, \textit{Senior Member, IEEE}
\IEEEcompsocitemizethanks{
\IEEEcompsocthanksitem Y.-H. Wu and M.M. Cheng
  are with the TKLNDST, College of Computer Science, Nankai University,
  Tianjin 300350, China. (E-mail: wuyuhuan@mail.nankai.edu.cn, cmm@nankai.edu.cn)
\IEEEcompsocthanksitem Y. Liu is with ETH Zurich. (E-mail: yun.liu@vision.ee.ethz.ch)
\IEEEcompsocthanksitem L. Zhang is with the School of Information and 
  Communication Engineering, 
  University of Electronic Science and Technology of China. (E-mail: zhangleuestc@gmail.com)
\IEEEcompsocthanksitem W. Gao is with the Science and Technology on 
  Complex System  Control  and  Intelligent  Agent  Cooperation Laboratory, 
  Beijing, China. (E-mail: gaowang\_fly@163.com)
\IEEEcompsocthanksitem M.-M. Cheng (cmm@nankai.edu.cn) and 
  W. Gao (gaowang\_fly@163.com) are corresponding authors.
}
}

\markboth{IEEE Transactions on Image Processing,~Vol.~30, Mar. 2021}%
{Wu \MakeLowercase{\textit{et al.}}: Regularized Densely-connected Pyramid Network for Salient Instance Segmentation}

\maketitle

\begin{abstract}
Much of the recent efforts on salient object detection (SOD)
have been devoted to producing accurate saliency maps 
without being aware of their instance labels. 
To this end, we propose a new pipeline for end-to-end salient instance 
segmentation (SIS) that predicts a class-agnostic mask for 
each detected salient instance. 
To better use the rich feature hierarchies in deep networks
and enhance the side predictions, 
we propose the regularized dense connections, 
which attentively promote informative features and 
suppress non-informative ones from all feature pyramids.
A novel multi-level RoIAlign based decoder is introduced to 
adaptively aggregate multi-level features for better mask predictions. 
Such strategies can be well-encapsulated into the Mask R-CNN pipeline. 
Extensive experiments on popular benchmarks demonstrate that 
our design significantly outperforms existing \sArt competitors by 
6.3\% (58.6\% vs. 52.3\%) in terms of the AP metric.
The code is available at \url{https://github.com/yuhuan-wu/RDPNet}.

\end{abstract}

\begin{IEEEkeywords}
salient instance segmentation, feature pyramid, RoIAlign\end{IEEEkeywords}

\IEEEpeerreviewmaketitle

\section{Introduction}

\IEEEPARstart{A}{s} a fundamental image understanding technique, 
salient object detection (SOD) aims at segmenting the most eye-attracting 
objects in a natural image.
Although recent SOD approaches 
\cite{zhang2017amulet,wang2017salient,wang2018salient,liu2019pool,liu2019dna} 
have achieved much success,
their generated saliency maps cannot discriminate different salient instances, 
which has prevented many applications from applying SOD for 
instance-level image understanding \cite{fan2018associating}.
Motivated by \cite{li2017instance}, in this paper, 
we tackle the more challenging case of SOD, 
called salient instance segmentation (SIS).
SIS segments salient objects from an image and discriminates salient instances 
by associating each instance with a different label.
SIS can facilitate more advanced tasks than SOD, 
such as image captioning \cite{fang2015captions}, 
weakly-supervised semantic/instance segmentation 
\cite{liu2020leveraging,21SC_WebSeg}, 
and visual tracking \cite{hong2015online}.

The MSRNet~\cite{li2017instance} made the first attempt to detect 
salient instances by adopting several isolated processing steps.
However, its performance was usually limited in challenging scenarios 
because it was not end-to-end trainable.
The S4Net \cite{fan2019s4net} replaced \emph{RoIAlign} in Mask R-CNN 
\cite{he2017mask} with the \emph{RoIMasking} to keep the scale of 
the feature maps and leverage the nearby background of objects.
Although much better performances were reported, 
it was far from satisfactory because only a limited feature level 
was utilized to decode salient instances.
One may argue that a natural solution is to employ the Feature Pyramid Network 
(FPN)~\cite{lin2017feature} and solve this task using the feature pyramid.
FPN builds the feature pyramid via the top-down pathway and lateral connections 
from the backbone.
With this network, small and large objects are more likely 
to be detected in the pyramid's low and high levels, respectively.
Therefore, apart from detecting the salient objects, 
much of the information flow was devoted to 
detecting the small and unnoticeable objects with the top-down pathway.
Naively applying the FPN architecture for SIS is suboptimal 
because salient objects are often much larger and distinctive than
noisy background and uninteresting objects.

\CheckRmv{
\begin{figure}[t!]
  \centering
  \includegraphics[width=\linewidth]{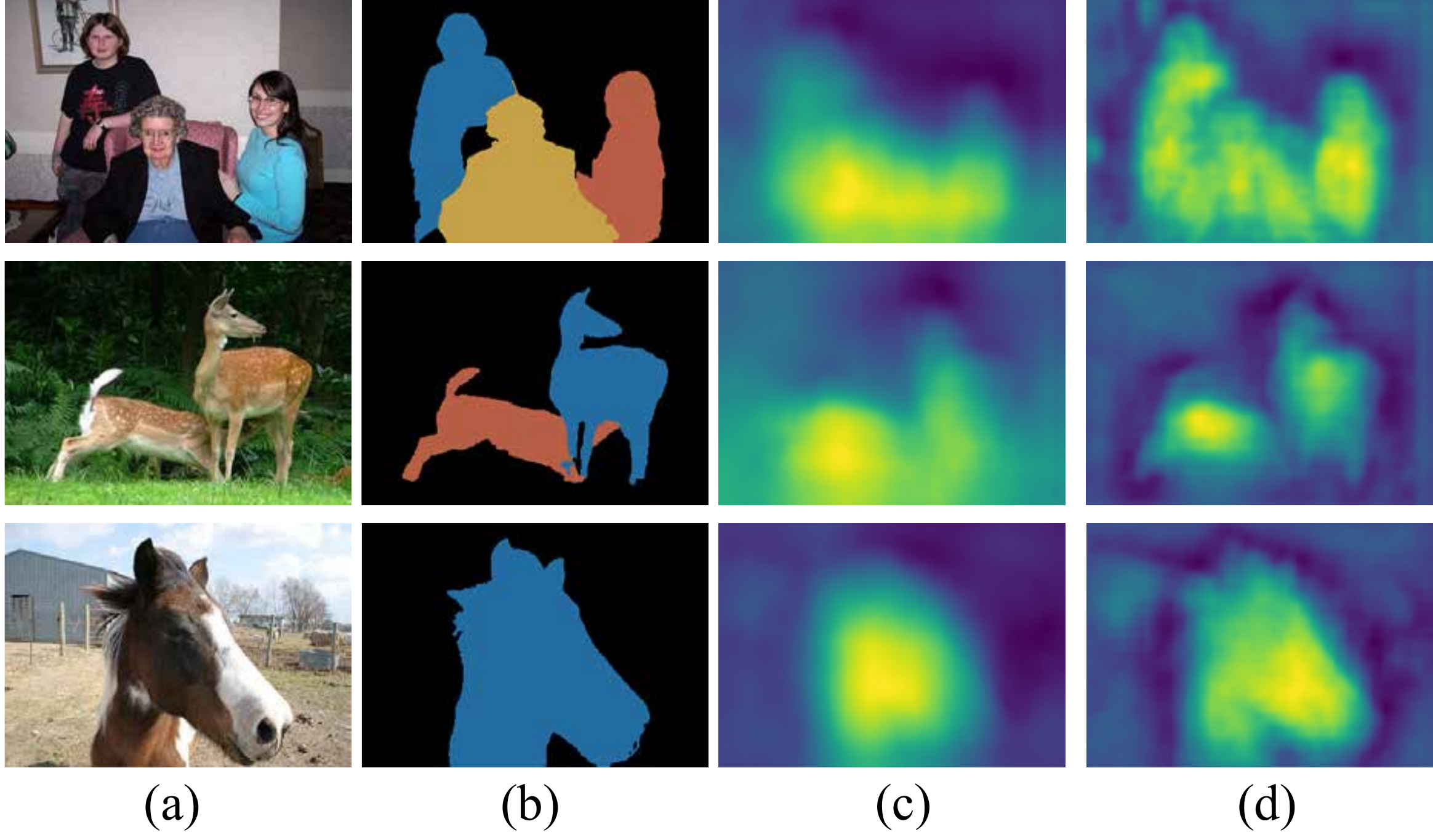}
  \vspace{-0.25in}
  \caption{Visualizations for the feature maps after passing FPN and 
    our proposed regularized densely-connected pyramid (RDP).
    (a) Source images; 
    (b) Corresponding ground truth; 
    (c) Visualized maps for the feature maps after FPN; 
    (d) Visualized maps for the feature maps after the proposed RDP.
    The visualized feature maps directly obtained by the FPN look coarser 
    and hard to recognize objects. 
    With our proposed RDP, 
    it is much easier to recognize each salient instance's locations and shapes.
  }\label{fig:visualization}
\end{figure}
}

Motivated by this, we focus on enhancing the side predictions by 
providing each side branch with richer feature hierarchies 
from deep networks to locate the object and recover its details.
We achieve this by proposing the regularized densely-connected pyramid 
(RDP) network, 
which provides richer feature hierarchies for each branch with dense connections.
In this way, each level can leverage both high-level semantic and 
low-level fine-grained features.
However, directly leveraging dense connections may yield noisy predictions 
due to different receptive fields of features with different feature levels.  
To this end, we propose to regularize such dense connections 
using the attention mechanism to promote informative features and 
suppress non-informative ones from all feature levels of the feature pyramid.

Our effort starts with Mask R-CNN \cite{he2017mask} that first 
detects bounding boxes and then adopts \emph{RoIAlign} to 
predict the binary mask for each region of interest (RoI).
Specifically, we propose the regularized densely-connected pyramid (RDP) 
network mentioned above to better enhance the feature pyramid with different 
scales while keeping semantic features for detecting salient instances.
More specifically, each level of features will be fused with not only 
its successive bottom features, as done in other works
\cite{lin2017feature,lin2017focal,wu2020jcs,fan2019s4net,liu2020rethink}, 
but also features from all the lower levels.
The RDP network only costs 0.7ms, 
which can be ignored in affecting the speed of the whole network.
\figref{fig:visualization} shows the superiority of RDP 
in feature learning compared with the FPN.
Besides, for mask prediction, 
traditional strategies like Mask R-CNN only use a specific feature level.
Which level is used is determined by the size of objects.
This design is suboptimal for SIS, 
and leveraging all feature levels is a better strategy.
Motivated by this, we propose to leverage the feature maps from 
all feature levels with a novel multi-level \emph{RoIAlign} operation
for extracting hierarchical RoIs for better mask prediction.
Extensive experiments demonstrate that the proposed method achieves \sArt 
performance and far surpasses previous competitors in terms of all metrics.
With an NVIDIA TITIAN Xp GPU, the proposed method runs at 45.0fps
for images of the $\sim320\times 480$ size and is thus suitable for 
real-time applications.

Overall, our main contributions are summarized as below:
\begin{itemize}
\item We propose regularized dense connections to attentively 
  promote informative features and suppress non-informative ones 
  at each stage of the feature pyramid,  
  providing richer bottom-up information flows.
\item We further propose a novel multi-level RoIAlign based decoder 
  to pool multi-level features for better mask predictions adaptively.
\item We empirically evaluate the proposed method on two popular SIS datasets 
  and demonstrate its superior accuracy and better efficiency.
\end{itemize}

\section{Related Work}

\subsection{Salient Object Detection}

SOD aims to detect salient objects or regions in natural images.
Conventional methods \cite{achanta2009frequency,cheng2013efficient,
cheng2014global,yang2013saliency,wang2017salient}
mainly focus on designing hand-crafted features and better prior strategies 
for SOD.
Later, some learning-based features \cite{wang2017salient} were studied as well.
Due to their limited representational ability, 
these methods have been suppressed by the deep learning-based methods.
Motivated by the success of convolutional neural networks (CNNs) and 
fully convolutional networks (FCNs) \cite{long2015fully},
many FCN-based SOD networks were proposed 
\cite{liu2016dhsnet,zhang2017amulet,wang2018salient,hou2019deeply,
liu2019dna,qiu2020simple,wu2020edn,liu2019pool,wu2020mobilesal,
gao2020highly,MINet-CVPR2020,liu2020lightweight,zhang2019salient}.
For example, Wang \etal \cite{wang2018salient} developed 
a recurrent FCN architecture for saliency prediction.
Liu \etal \cite{liu2016dhsnet} presented a deep hierarchical saliency network 
to learn a coarse global prediction and refine it hierarchically 
and progressively by integrating local information.
Inspired by \cite{xie2017holistically,liu2019richer}, 
Hou \etal \cite{hou2019deeply} introduced short connections for side-outputs 
to enrich multi-scale features.
Zhang \etal \cite{zhang2017amulet} introduced a bi-directional structure 
to adaptively aggregate multi-level features.
Wang \etal \cite{wang2018detect} proposed to detect salient objects globally 
and recurrently refine the saliency maps.
Liu \etal \cite{liu2018picanet} proposed a pixel-wise contextual attention 
network to selectively attend to each pixel's informative context locations.
Liu \etal \cite{liu2019pool} proposed various pooling-based modules 
to strengthen the feature representations with real-time speed.
More details of the development in SOD can refer to 
\cite{borji2019salient, borji2015salient, wang2019focal, wangsalient}.
Although these methods can detect saliency maps accurately, 
they cannot discriminate different salient object instances.

\subsection{Instance Segmentation}
%
Similar to object detection, early instance segmentation works \cite{girshick2014rich,hariharan2015hypercolumns,dai2015convolutional}
focus on classifying segmented proposals generated by object proposal methods 
\cite{uijlings2013selective,pont2017multiscale,liu2020refinedbox,cheng2019bing}.
Li \etal \cite{li2017fully} first proposed an end-to-end 
fully convolutional instance segmentation (FCIS) framework.
He \etal \cite{he2017mask} extended Faster R-CNN \cite{ren2015faster} to 
Mask R-CNN by replacing \emph{RoIPool} with \emph{RoIAlign} 
for more accurate RoI generation.
They added a parallel mask head with the box head in Faster R-CNN 
for mask prediction using the feature pyramid's RoI features.
Mask scoring (MS) R-CNN~\cite{huang2019mask} combines the 
mask confidence score and the localization score and 
is thus more precise for scoring the detected instances.
HTC \cite{chen2019hybrid} proposes a hybrid multi-stage cascade for 
both box and mask detection.
Based on FCOS \cite{tian2019fcos}, 
CenterMask \cite{lee2020centermask} designs spatial-attention-guided 
mask prediction for anchor-free instance segmentation.
BlendMask \cite{chen2020blendmask} achieves instance segmentation 
via a blender with the learned bases and instance attentions.
SOLO \cite{wang2020solo} segments object for each location.
DetectoRS \cite{qiao2020detectors} proposes the recursive feature pyramid and
switchable atrous convolution for better performance.

\subsection{Feature Pyramid Enhancement}

The feature pyramid is known as a powerful tool for strengthening multi-scale
feature representations \cite{felzenszwalb2008discriminatively}.
The necessity of feature pyramid enhancement has also been demonstrated 
in detecting locations \cite{lin2017feature} or 
segmenting objects \cite{ronneberger2015u}.
The early successor FPN \cite{lin2017feature} builds the feature pyramid via 
the top-down pathway and lateral connections from the backbone feature pyramid.
PANet \cite{liu2018path} builds upon FPN and adds 
an extra bottom-up path augmentation.
NAS-FPN \cite{ghiasi2019fpn} extends the idea of FPN by learning the
scalable feature pyramid architecture using neural architecture search (NAS).
EfficientDet \cite{tan2020efficientdet} proposes BiFPN, which optimizes
multi-scale feature fusion in a more efficient bidirectional manner.

\subsection{Salient Instance Segmentation}
SIS is a relatively new problem that shares similar spirits with both SOD 
and instance segmentation.
It is more challenging than SOD because it segments salient objects 
and meanwhile differentiates different salient instances.
One possible solution is to derive the salient instances directly from 
the saliency map using some post-processing techniques.
For example, Li \etal \cite{li2017instance} proposed a two-stage solution, 
called MSRNet, which first produces saliency maps and salient object contours 
that are then integrated with MCG \cite{pont2017multiscale} for SIS.
Although MSRNet can learn from the saliency maps, as the two stages 
are optimized isolatedly, the results are not satisfactory.
To overcome the isolated optimization difficulties, recently, 
Fan \etal \cite{fan2019s4net} introduced an end-to-end single-stage framework 
based on the Mask R-CNN \cite{he2017mask}.
They learned to mimic the strategy of GrabCut \cite{rother2004grabcut} 
and used the so-called \emph{RoIMasking} to incorporate 
foreground/background separation explicitly.
They also designed a customized segmentation head with dilated convolutions 
to retrieve instance masks from the coarsest feature level.
Instead of using a single specific feature level with limited semantic features 
as done in existing methods, 
we propose to use the regularized densely-connected pyramid (RDP) networks
to extract 
richer feature hierarchies with higher contrasts
(as in \figref{fig:visualization}) from all feature levels. 
Our design significantly releases the burden of accurately detecting 
salient instances and retrieving binary masks for each salient instance.

\section{Our Approach}


\subsection{Feature Pyramid Enhancement}\label{sec:feature_enhance}
%
The feature pyramid, which is usually understood as a group of feature maps 
with different resolutions, 
has demonstrated its superiority in various computer vision tasks.
One notable application is object detection, 
which aims to detect semantic objects' locations accurately.
As there are large-scale variations for natural objects, 
directly detecting targets' accurate locations by simply using features 
from one scale is extremely challenging. 
Therefore, 
many researchers attempt to detect semantic objects with the feature pyramid.
Our method naturally belongs to this family.
We propose a densely-connected pyramid (DP) network and the advanced regularized densely-connected pyramid (RDP) network for the feature pyramid enhancement.
We elaborate on the main idea below.


\subsubsection{Problem Formulation}
Given an image as the input and a base network (\eg, ResNet \cite{he2016deep}) for feature extraction, we can first derive a set of side-outputs from multiple stages in this network.
Assume that we have access to multiple scales of features $\{C_m, C_{m+1}, \cdots, C_k\}$ from the $m$-th to $k$-th stage, corresponding to the finest and coarsest feature maps.
Typically, $m$ will be 2 as defined in two-stage detectors like Faster R-CNN \cite{ren2015faster,lin2017feature} or 3 as defined in one-stage detectors like RetinaNet \cite{lin2017focal}.
$k$ is typically 5 as defined in both kinds of detectors~\cite{ren2015faster,lin2017feature,lin2017focal}.

\subsubsection{The Top-down Style}
%
%
In order to leverage both high-level semantics and low-level fine details as mentioned above, the well-known FPN \cite{lin2017feature} proposes a top-down architecture with lateral connections to strengthen the capacity and representability of each side-output.
Such a strategy has been demonstrated very powerful especially for detecting small and tiny objects and has been extensively used in many other approaches.
Suppose that the feature pyramid enhanced by FPN is called $P=\{P_m, P_{m+1}, ..., P_k\}$.
This enhancement operation can be formulated as:
%

\CheckRmv{
\begin{equation}
P_{k} = \mathcal{F}[\phi(C_k)],
\label{fpn_oneside}
\end{equation}
\begin{equation}
P_{i} = \mathcal{F}[\phi(C_i) + Upsample(P_{i+1})], m \le i < k,
\label{equ:fpn}
\end{equation}
}
where $\phi$ represents a $1\times 1$ convolution layer to reduce the channels of $C_i$.
$\mathcal{F}$ represents the feature fusion module which consists of a single $3\times 3$ convolution layer.
The upsampling factor for $P_{i+1}$ is 2 and we use the bilinear interpolation for upsampling.
For the coarsest feature map $C_k$, this enhancement operation is simplified done by passing a single $3\times 3$ convolution.


Such a strategy, however, is suboptimal for SIS.
Recall that the objective of this task is to detect salient instances and ignore other non-salient ones that usually have a relatively smaller size.  
In~\equref{equ:fpn}, each side branch only has limited bottom-up information because it only leverages the features of two successive layers.
In this way, higher levels in the pyramid have limited access to the low-level fine-grained details and thus may fail to recover the instance boundaries.
Similarly, the lower levels in the pyramid lack the high-level semantic information and thus may not be good at accurately locating the salient objects and identifying their instance labels.
To address this problem, we provide our solution below.


\CheckRmv{
\begin{figure}[t!]
    \centering
    \includegraphics[width=.82\linewidth]{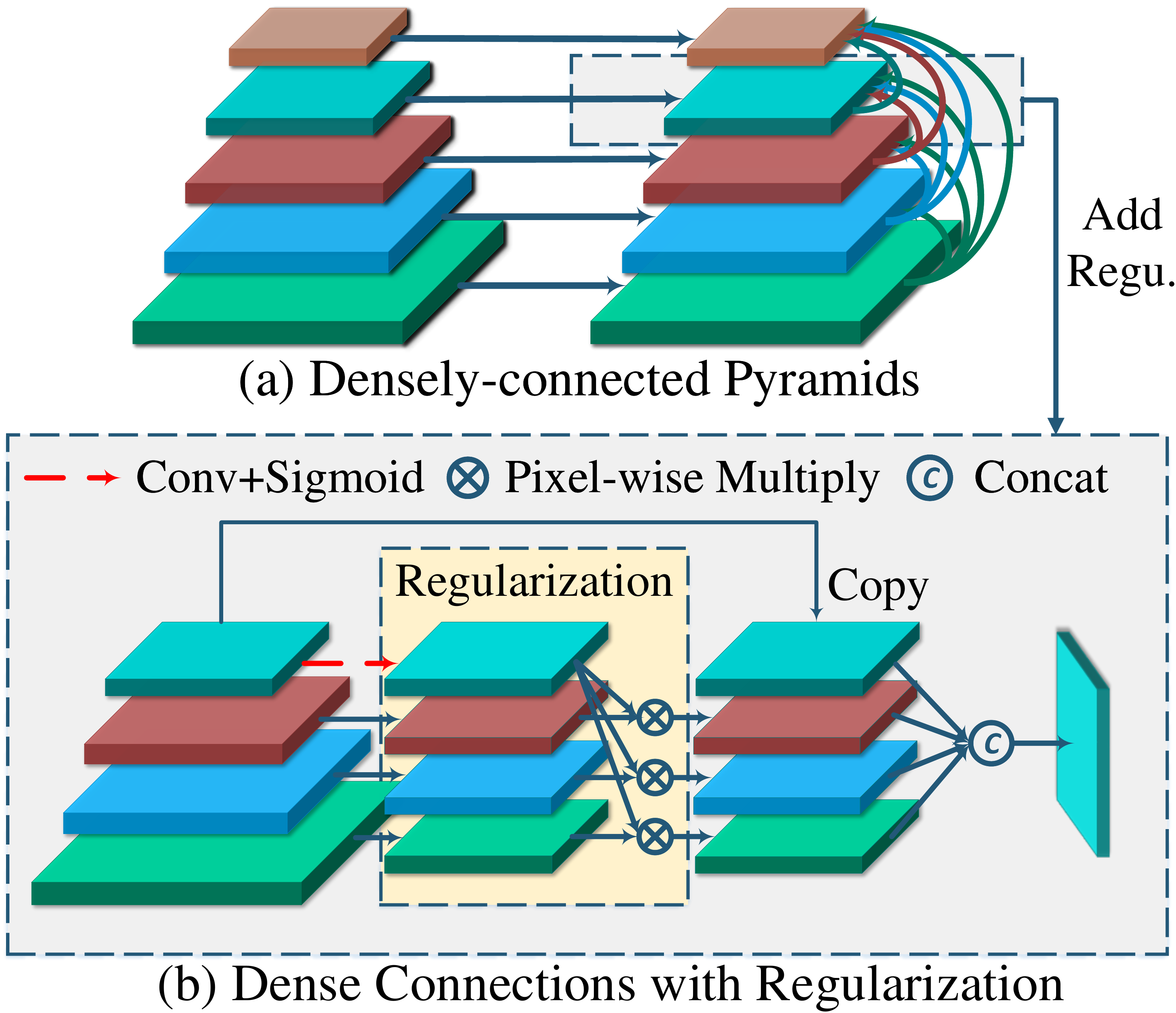}
    \vspace{-0.12in}
    \caption{Illustration of the proposed regularized densely-connected pyramid (RDP) network for feature pyramid enhancement.
    (a) The densely-connected pyramid (DP) network;
    (b) Dense connections with regularization.
    For simplicity, we illustrate the regularization with only the $4^{th}$ feature level.
    RDP is DP with the regularization at each feature level.
    } \label{fig:rdp}
\end{figure}
}

\subsubsection{The Bottom-up Densely-connected Pyramid Network}
%
%
A straightforward solution to overcome the above-mentioned disadvantages of FPN, as proposed in \cite{liu2018path}, is to build a progressive bottom-up lateral connection and recreate a new feature pyramid:
\CheckRmv{
\begin{equation}
P_m' = \mathcal{F}(P_m),
\end{equation}
}
\CheckRmv{
\begin{equation}
P_{j+1}' = \mathcal{F}[P_{j+1} + Downsample(P_{j}')], m < j \le k-1,
\end{equation}
}
where $P_k'$ is the re-generated feature map of the new feature pyramid.
This solution naturally follows a progressive manner of
the FPN and is applied in instance segmentation \cite{liu2018path}.
We take inspiration from this architecture and make necessary amendments.
For each feature level in the network, instead of only merging two successive levels, we merge features from many other levels as well.
This design is advantageous because each stage is given a much richer information flow from all its bottom layers.
More specifically, we achieve this by adding dense connections, which can be formulated as
\CheckRmv{
\begin{equation}
P_j' = \mathcal{F}\{\phi[Concat(P_m',P_{m+1}',...,P_{j-1}', P_j)]\},
\end{equation}
}
where we have $m < j \le k$ and $m$ represents the index of the first stage of the feature pyramid.
In the concatenation operation, feature maps $P_m', P_{m+1}', ..., P_{j-1}'$ are all downsampled to the size of $P_j$.
We use the $1\times 1$ convolution operation $\phi$ to reduce the channels to that of $P_j$.


\CheckRmv{
\begin{figure*}[t!]
    \centering
    \includegraphics[width=.9\linewidth]{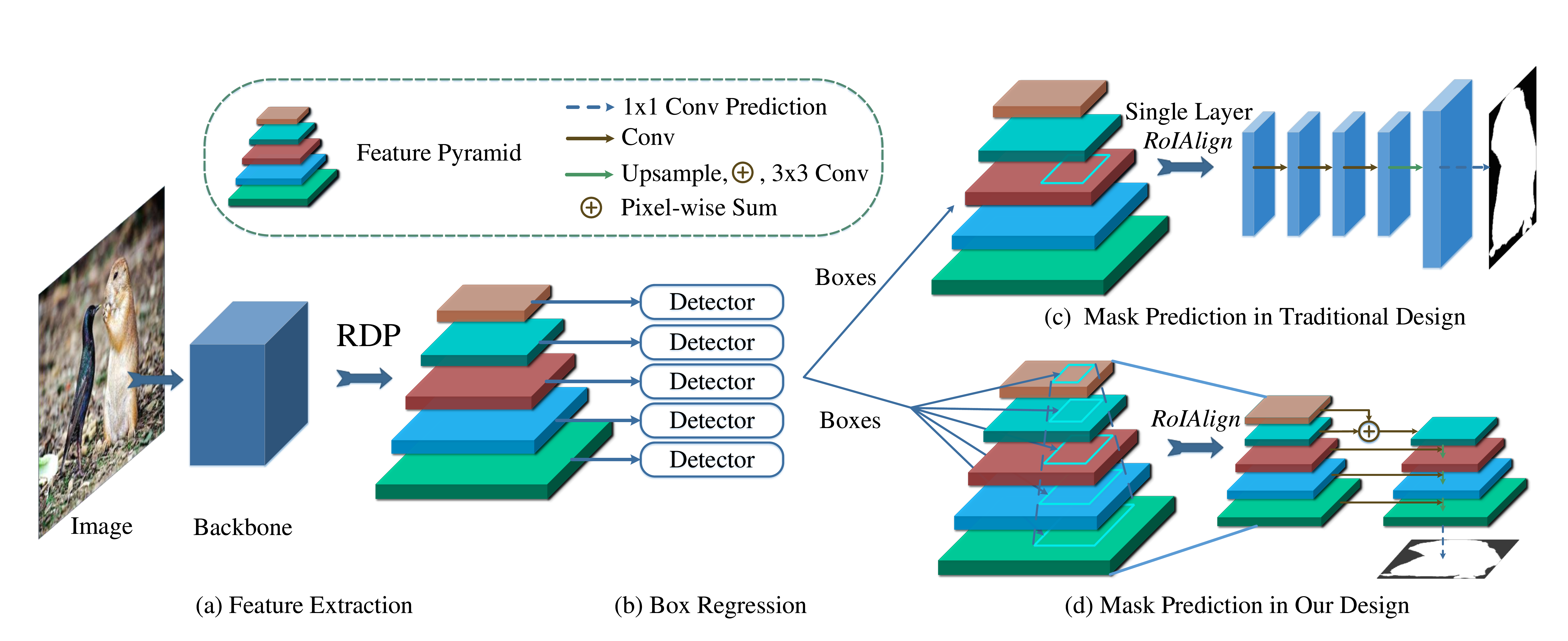}
    \caption{The overall pipeline of the proposed method.
    (a) In the feature extraction part, RDP is the regularized densely-connected pyramid network, as illustrated in \figref{fig:rdp}.
    (b) We use the base detector \cite{tian2019fcos} for box regression at each feature level.
    (c) The traditional design for mask prediction only uses a single layer to decode the binary masks.
    (d) Our design for mask prediction uses all feature levels to decode binary masks by a simple decoder.
    } \label{fig:pipeline}
\end{figure*}
}

\subsubsection{Regularized Densely-Connected Pyramid Network} \label{sec:rdp}
The bottom-up dense connections essentially expand the input space for each side branch.
However, as features from different layers usually have different receptive fields, they are usually not very compatible in discovering the fine details of the object due to the scale conflict.
To this end, we further regularize the dense connections with the well-established self-attention mechanism.
To compute the new feature maps $P_j'$, we first create spatial regularization based on the feature map $P_j$ of the current scale:
\CheckRmv{
\begin{equation}
R_j=\sigma \mathcal{F}(P_j),
\end{equation}
}
where $R_j$ is the attention map for the regularization and $\sigma$ denotes the sigmoid function for each pixel.
By reducing the effect of the scale conflict during the feature concatenation, we apply this regularization
to the feature maps with identical attention maps $R_j$ in the feature fusion except for $P_{j}$:
\CheckRmv{
\begin{equation}
P_{t}^r = R_j \otimes Downsample(P_{t}'), m \le t < j,
\end{equation}
}
where $P_t^r$ is the regularized feature map from other scales.
We perform the downsampling operation to features maps from other scales of the same size as $P_j$.
The symbol $\otimes$ denotes the element-wise multiplication.
Overall, the regularized dense connections for enhancing the feature pyramid can be formulated as
\CheckRmv{
\begin{equation}
P_j' = F[Concat(P_m^r,P_{m+1}^r,..., P_{j-1}^r,P_j)], m < j \le k.
\end{equation}
}
We provide an illustration of the proposed RDP in \figref{fig:rdp} for better understanding.

\subsection{Multi-level RoIAlign for Mask Prediction}
\label{sec:multi_align}
%
Mask prediction is essential for SIS as it directly determines the accuracy of the mask for each salient instance.
As shown in \figref{fig:pipeline} (c), Mask R-CNN \cite{he2017mask} uses a specific feature level, which depends on the size of the object of interest, for mask prediction using \emph{RoIAlign}.
Although this option of determining which feature level is used in \emph{RoIAlign} can adaptively extract masks for objects of different sizes, this is suboptimal for SIS and a better strategy is to leverage all the feature levels.
More specifically, we propose an efficient yet well-performing multi-level \emph{RoIAlign} with a decoder to leverage all feature levels.
\figref{fig:pipeline} (d) illustrates our idea.
After the multi-level \emph{RoIAlign} layer, we derive a tiny feature pyramid specifically for the mask prediction.
The next decoder is to progressively decode the binary masks from the tiny feature pyramid.
The decoder consists of the lateral connections and some feature fusion operations.
Since the strides of the top two feature maps are very large, they are \emph{RoIAligned} to the same size of RoIs, and we perform element-wise sum for these two RoIs.
Other feature maps are \emph{RoIAligned} to different sizes of RoIs.

With this decoder, we first use \emph{RoIAlign} to adaptively align features from all levels 
and then retrieve binary masks based on the aligned features.
For the feature fusion between two adjacent feature maps of different sizes, we first perform bilinear interpolation to upsample them to the size of the finer feature map by a factor of 2.
Then, we use the element-wise sum to fuse these two feature maps 
and add a $3\times 3$ convolution layer to generate the new feature maps for the next feature fusion.
Finally, we get the finest feature maps, on which we perform a $1\times 1$ convolution to predict the binary masks.

\subsection{Overall Pipeline}\label{sec:pipeline}
The regularized densely-connected pyramid and the multi-level \emph{RoIAlign} layer are encapsulated into a Mask R-CNN based pipeline, as displayed in \figref{fig:pipeline}.
The functionality of each component is presented in the following.

\subsubsection{Feature Extraction}
We adopt the widely used ResNet \cite{he2016deep} as our backbone network, which has been pretrained on the ImageNet dataset \cite{russakovsky2015imagenet}.
The base feature pyramid follows the architecture of FPN \cite{lin2017feature}.
Since we use the one-stage detector~\cite{tian2019fcos} for box regression, we follow \cite{tian2019fcos} to generate two extra feature maps, $P_{6}$ and $P_{7}$, by connecting two $3\times 3$ convolutions with a stride of 2 after $P_5$.
$P_{6}$ and $P_{7}$ are added to the feature pyramid, so the feature pyramid after passing FPN is $\{P_3,P_4,P_5,P_6,P_7\}$.
All feature maps in this feature pyramid are with 256 channels.
Then, we build the regularized densely-connected pyramid (RDP) from $P_3'$ to $P_7'$, as introduced in \secref{sec:rdp} and \figref{fig:rdp}.
The number of output channels is still 256 for all feature maps in the reconstructed feature pyramid.
\figref{fig:visualization} displays the visualization of feature maps after passing FPN and our proposed RDP.
We find that although feature maps derived by FPN have captured the locations of salient instances, the activation or high responses are very coarse or cannot even recognize the number of salient instances in each image.
In contrast, the feature maps from our proposed RDP network have more precise activation and can help the base detector better to detect the bounding box of each salient instance.
This design further enhances the mask head towards obtaining better masks for the detected salient instances.

\subsubsection{Box Regression} \label{sec:box_reg}
To quickly detect the salient instances, we do not apply a heavy two-stage detector 
that contains an RPN \cite{ren2015faster} head to generate object proposals 
and classifies these object proposals with the box head
because it is too slow for SIS.
Instead, we use the one-stage detector FCOS \cite{tian2019fcos} as our base detector.
This detector consists of four convolution-ReLU layers with 256 channels, and the box regression is performed at each feature level with this shared-parameters head.
The details for calculating the box proposals from the final feature map can refer to \cite{tian2019fcos}.
In this part, we will derive many box proposals with their confidence scores in each feature level.
We concatenate them and leave the top 1000 boxes with a confidence score larger than 0.05.
After that, a non-maximum suppression (NMS) operation is conducted on the boxes and then keep at most top 100 boxes for predicting their corresponding binary masks.

\subsubsection{Mask Prediction}
In the box regression, we detect the salient instances in the box level.
Since our final goal is to predict the instance-level segmentation, mask prediction is necessary to retrieve the corresponding binary mask for each salient instance.
We make a further improvement to Mask R-CNN by leveraging the feature maps of all feature levels ($\{P_3',P_4',P_5',P_6',P_7'\}$) for retrieving binary masks for salient instances.
After the multi-level \emph{RoIAlign} layer, the sizes of the feature maps $\{D_3,D_4,D_5,D_6,D_7\}$ are displayed in \tabref{tab:size_rois}.
Please refer to \secref{sec:multi_align} for the implementation of the decoder.
After passing this decoder, we use a simple $1\times 1$ convolution layer to predict the final masks for the detected salient instances.

\CheckRmv{
\begin{table}[htbp]
    \centering
    \caption{Feature map size for each channel after the multi-level \emph{RoIAlign} layers. Since $P_7'$ and $P_6'$ are very small, $D_7$ and $D_6$ are sampled with the same size. The size of the final mask for each salient instance is $32\times 32$.}
    \label{tab:size_rois}
    \begin{tabular}{c|ccccc} \hline
    Name  & $D_7$ & $D_6$ & $D_5$ & $D_4$ & $D_3$ \\ \hline
    Size & $4 \times 4$ & $4 \times 4$ & $8 \times 8$ & $16 \times 16$ & $32 \times 32$ \\ \hline
    \end{tabular}
    \vspace{1mm}
\end{table}
}

\subsubsection{The Loss Function}
%
Our pipeline has two key parts that need supervisions: box regression and mask prediction.
A foreground box classification loss $L_{cls}$ and a coordinate regression loss $L_{reg}$ are applied in the box regression branch.
Note that $L_{cls}$ is the focal loss \cite{lin2017focal} and $L_{reg}$ is the IoU loss proposed in \cite{yu2016unitbox}.
To further get rid of the bad effect of too many low-quality boxes, we apply the centerness loss $L_{center}$ proposed in \cite{tian2019fcos} to ignore the boxes whose centers are far away from the centers of salient instances.
For mask prediction, we use the standard cross-entropy loss as the mask loss $L_{mask}$.
Hence we obtain the final loss $L = L_{cls} + L_{reg} + L_{center} + L_{mask}$ to supervise the whole network.

\section{Experiments}
In this section, we will first introduce the datasets and evaluation metrics used in our experiments, as in \secref{dataset}.
Implementation details will be described in \secref{implementaion_detail}.
We will carefully examine our proposed designs and demonstrate their effectiveness in \secref{ablation}.
The results of our method and the comparison with previous \sArt methods will be provided in \secref{comparison}.

\subsection{Dataset and Evaluation Metric}\label{dataset}
\subsubsection{Datasets}
We adopt two popular datasets in our experiments, \ie, ISOD and SOC datasets.
The ISOD dataset is proposed by Li \etal \cite{li2017instance}.
It contains 1000 images with salient instance annotations.
We follow the previous work \cite{fan2019s4net} to use 500 images for training, 200 images for validation, and another 300 images for testing.
The SOC dataset is proposed by Fan \etal \cite{fan2018salient}.
This dataset consists of 3000 images in cluttered scenes with salient instance annotations.
Among them, 2400 images are used for training and the other 600 images are used for testing.

\subsubsection{Evaluation Metrics}
Previous works use the mAP metric with a specific threshold such as 0.5 (standard) or 0.7 (strict) to determine whether a detected instance is a true positive (TP), similar to the evaluation in the PASCAL VOC challenge \cite{everingham2010pascal}.
However, as this metric is not enough to fully reflect the quality of detectors, the MS-COCO evaluation metric \cite{lin2014microsoft} has been widely used in mainstream object detection and instance segmentation.
We follow the MS-COCO evaluation metric \cite{lin2014microsoft} 
to use mAP@\{0.5:0.05:0.95\} as the primary metric, 
since it can better reflect the detection quality.
We also report mAP@0.5 and mAP@0.7 for reference, as done in related works \cite{liu2016ssd,lin2017feature,lin2017focal,he2017mask,huang2019mask}.
For simplicity, we use ``AP'', ``AP$_{50}$'', and ``AP$_{70}$'' to stand for mAP@\{0.5:0.05:0.95\}, mAP@0.5, and mAP@0.7, respectively.

\subsection{Implementation Details}\label{implementaion_detail}
In this paper, we use the popular PyTorch \cite{paszke2019pytorch} 
and Jittor \cite{hu2020jittor} framework 
to implement our method.
If not specially mentioned, we apply the widely used ResNet-50 \cite{he2016deep} as the backbone network.
In the network training, maybe there is no box satisfying the threshold of the confidence score for NMS, especially in the early training stage, so we add the ground-truth boxes to the results of detected salient instances in the training to prevent such a situation to take place.
We only use horizontal flipping as the data augmentation, and each input image is resized as the shorter side is 320 pixels and the longer side follows the initial image aspect ratio but is limited to a maximum value of 480 pixels.
We use a single NVIDIA TITAN Xp GPU for all experiments.
We use the SGD optimizer with the weight decay of $10^{-4}$ and the momentum of 0.9.
Each mini-batch contains four images.
The initial learning rate is 0.0025.
For the ISOD dataset~\cite{li2017instance}, the learning rate is divided by 10 after 6K iterations, and we train our network for 9K iterations in total.
For the SOC dataset~\cite{fan2018salient}, the learning rate is divided by 10 after 24K iterations, and we train our network for 36K iterations in total.
Due to the small batch size, all the BatchNorm layers of the backbone network are frozen during training.
The $3\times 3$ convolution layers of the box regression head and mask prediction head are with the group normalization \cite{wu2018group}.
The number of output channels of each $3\times 3$ convolution layer is 128 in the mask prediction head.

\CheckRmv{
\begin{table}[!tb]
    \centering
    \caption{Evaluation on the ISOD validation set for various design choices.
        The first line refers to the baseline of FPN.
    	NP is the natural progressive bottom-up style for building the new feature pyramid.
    	DP denotes the proposed method that rebuilds the feature pyramid with dense connections.
    	RDP means to add the proposed regularization to DP.
    	MRA represents the proposed multi-level \textit{RoIAlign}.
    } \label{tab:ablation_ours}
    \begin{tabular}{ccc|ccc} \hline
          DP & RDP & MRA & AP & AP$_{50}$ & AP$_{70}$ \\ \hline\hline
         - & - & - & 54.2\% & 83.3\% & 69.7\% \\
        \ding{52} &  &  & 55.1\% & 84.4\% & 71.0\% \\
         &  & \revise{\ding{52}} & \revise{56.3\%} & \revise{85.5\%} & \revise{71.2\%} \\
         & \ding{52} &  & 56.4\% & 85.4\% & 72.0\% \\
         & \ding{52} & \ding{52} & \textbf{57.4\%} & \textbf{86.1\%} & \textbf{73.8\%} \\ \hline
    \end{tabular}
\end{table}
}

\subsection{Ablation Study} \label{ablation}
In this part, we evaluate the effect of various designs on the ISOD dataset.
We use its training set for training and report results on its validation set.
If not mentioned, we use the ResNet-50 as the backbone for our network.

\subsubsection{Effect of DP and RDP}
As mentioned in \secref{sec:rdp}, 
we propose to create the RDP to fill the vacancy of the FPN.
Here, we view FPN as our baseline and evaluate four design choices:
i) NP, \ie, the naive progressive bottom-up style for building the new feature pyramid;
ii) DP, \ie, the proposed method that rebuilds the feature pyramid with dense connections;
iii) RDP, \ie, adding the proposed regularization to the dense connections in DP;
iv) MRA, \ie, the proposed multi-level \textit{RoIAlign}.
\tabref{tab:ablation_ours} shows the evaluation results on the ISOD validation set.
If we add DP without regularization, the metric of AP will be improved by 0.9\% compared with FPN.
When we add the regularization to DP, a relative 1.3\% improvement over DP is observed, indicating that regularization is vital for the proposed densely-connected pyramid.
Note that the proposed RDP is very efficient and only costs 0.7ms for a $320\times 480$ input image, making it have little effect on the speed of the whole network.

\subsubsection{Effect of Multi-level \textit{RoIAlign}}
The existing research usually predicts object masks using the mask head proposed by Mask R-CNN \cite{he2017mask}, which predicts masks from a specific feature level.
Instead, we propose a top-down progressive mask decoder to utilize all feature levels for object mask prediction, namely multi-level \textit{RoIAlign} (MRA).
The comparison between MRA and the traditional \textit{RoIAlign} can be found in \tabref{tab:ablation_ours}.
\revise{One can see that applying MRA on the baseline brings 2.1\%, 2.2\%, and 1.5\% improvement 
in terms of AP, AP$_{50}$, and AP$_{70}$, respectively.}
We can also observe that the introduction of MRA based on the baseline with the RDP further leads to an improvement of 1.0\%, 0.7\%, and 1.8\% in terms of AP, AP$_{50}$, and AP$_{70}$, respectively.
This demonstrates the significance of the proposed MRA in accurate mask prediction by leveraging all feature levels.
Overall, the proposed method achieves 3.2\% higher AP, 2.8\% higher AP$_{50}$, and 4.1\% higher AP$_{75}$ than the baseline of FPN.

\subsubsection{Partially Applying DP and RDP}
Our initial design considers all feature levels ($P_3$ - $P_7$) 
to reconstruct the feature pyramid.
Among them, the top 2 feature levels ($P_6$ and $P_7$) are generated from $P_5$ using only two $3\times 3$ convolutions.
In this section, we further evaluate the effectiveness of DP and RDP by applying them to a part of side-outputs.
Specifically, we only apply DP/RDP to three side-outputs, \ie, $P_3$, $P_4$, and $P_5$, excluding $P_6$ and $P_7$.
The experimental results are shown in \tabref{tab:less_levels}.
We could see that applying DP/RDP to only three side-outputs performs better than the baseline, but performs worse than applying DP/RDP to five side-outputs, indicating that DP/RDP is effective in feature enhancement for all feature levels.
The fact that RDP with only three feature levels significantly outperforms the baseline, further suggests that RDP is very useful to FPN.

\CheckRmv{
\begin{table}[!tb]
    \centering
    \caption{Evaluation on the ISOD validation set for partially applying DP/RDP to a part of side-outputs. $P_3 \sim P_5$ means from $P_3$ to $P_5$. $P_3 \sim P_7$ means all side-outputs in the feature pyramid.
    } \label{tab:less_levels}
    \begin{tabular}{ccc|ccc} \hline
        Side-outputs & DP & RDP & AP & AP$_{50}$ & AP$_{70}$ \\ \hline\hline
        - & -  & -& 54.2\% & 83.3\% & 69.7\% \\
        $P_3 \sim P_5$  & \ding{52}  &  & 54.4\% & 83.7\% & 70.3\% \\
        $P_3 \sim P_7$ & \ding{52} &  & 55.1\% & 84.4\% & 71.0\% \\
        $P_3 \sim P_5$ &  & \ding{52} & 55.8\% & 84.9\% & 71.3\% \\ %
        $P_3 \sim P_7$ &  & \ding{52} & 56.4\% & 85.4\% & 72.0\% \\ \hline
    \end{tabular}
\end{table}
}

\CheckRmv{
\begin{table}[!t]
    \centering
    \renewcommand{\tabcolsep}{4.0mm}
    \caption{Evaluation on the ISOD validation set for the top-down and bottom-up designs of RDP. The top-down design directly replace FPN of the baseline method with the top-down style of RDP. The bottom-up design is the default version of RDP as shown in \figref{fig:rdp}.
    } \label{tab:ablation_fail}
    \begin{tabular}{c|ccc} \hline
        Method & AP & AP$_{50}$ & AP$_{70}$ \\ \hline\hline
        Baseline & 54.2\% & 83.3\% & 69.7\% \\
        Top-down & 45.6\% & 76.9\% & 57.0\% \\
        Bottom-up & 56.4\% & 85.4\% & 72.0\% \\
    	 \hline
    \end{tabular}
\end{table}
}

\CheckRmv{
\begin{table}[!t]
    \centering
    \renewcommand{\tabcolsep}{4.0mm}
    \caption{\revise{Comparison of different feature pyramid enhancement strategies}.
    } \label{tab:fpn_cmp}
    \begin{tabular}{l|ccc} \hline
        Method & AP & AP$_{50}$ & AP$_{70}$ \\ \hline\hline
        Baseline & 54.2\% & 83.3\% & 69.7\% \\
        +PA \cite{liu2018path}  & 54.6\% & 84.9\% & 70.0\% \\
        +NAS-FPN \cite{ghiasi2019fpn} & 54.3\% & 84.6\% & 69.6\% \\
        +BiFPN \cite{tan2020efficientdet} & 54.1\% & 84.3\% & 69.7\% \\
        +RDP & \textbf{56.4\%} & \textbf{85.4\%} & \textbf{72.0\%} \\
    	 \hline
    \end{tabular}
\end{table}
}

\newcommand{\addFig}[1]{{\includegraphics[height=.19\textwidth]{curve/error_ana/#1.pdf}}}

\CheckRmv{
\begin{figure*}[!t]
\centering
\renewcommand{\arraystretch}{1.8}
\setlength{\tabcolsep}{1mm}
\begin{tabular}{cccc}
    \addFig{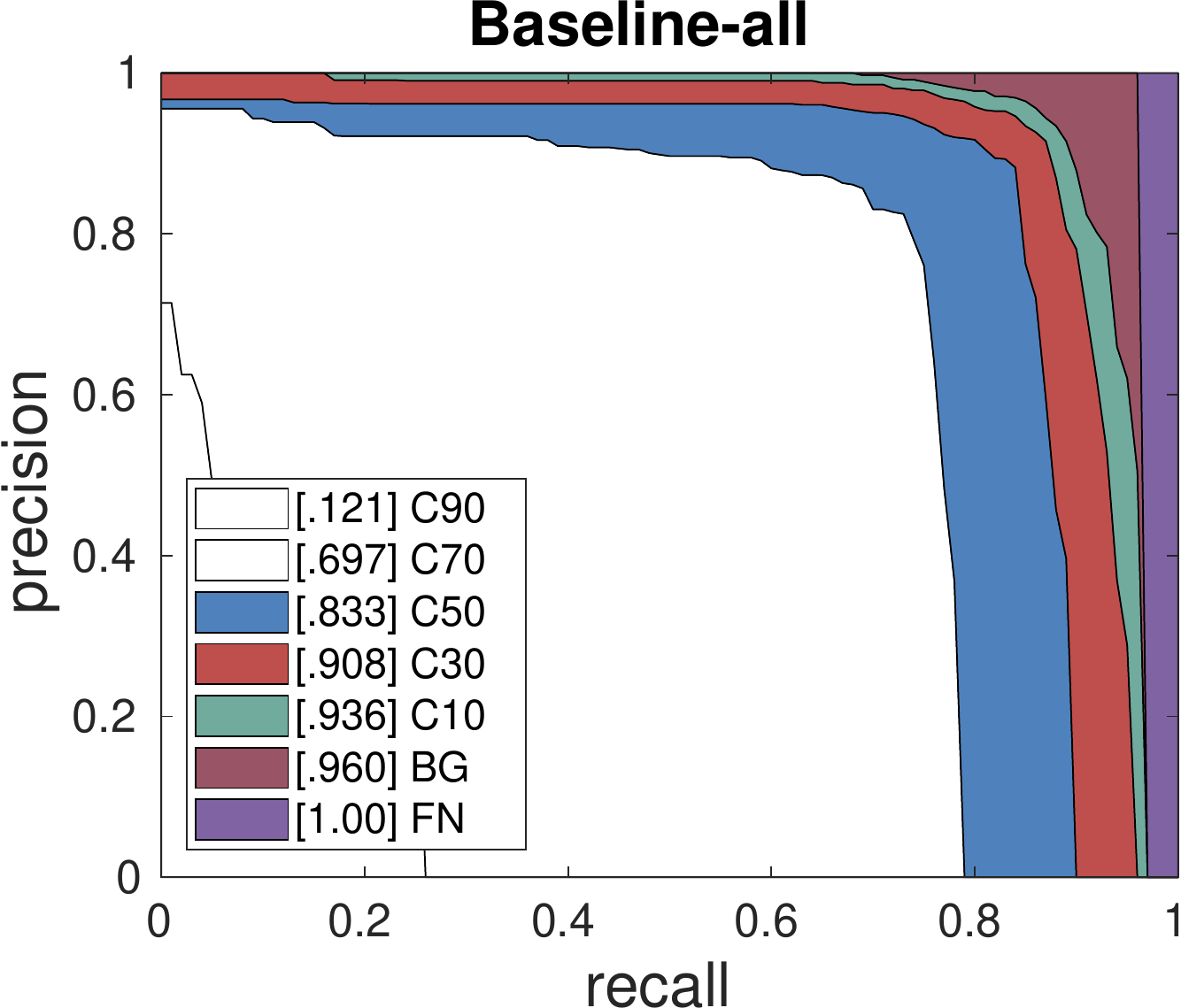} &
    \addFig{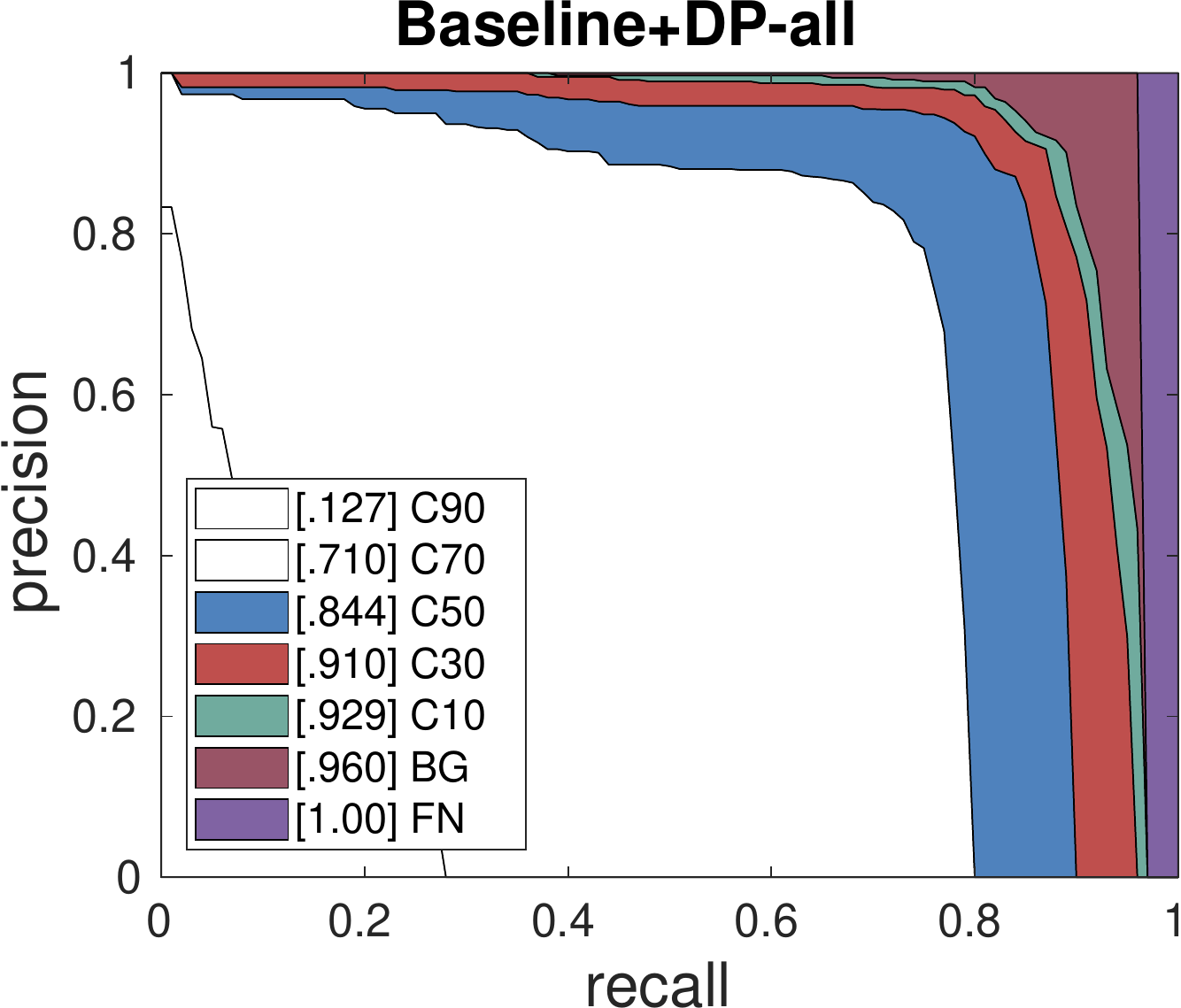} &
    \addFig{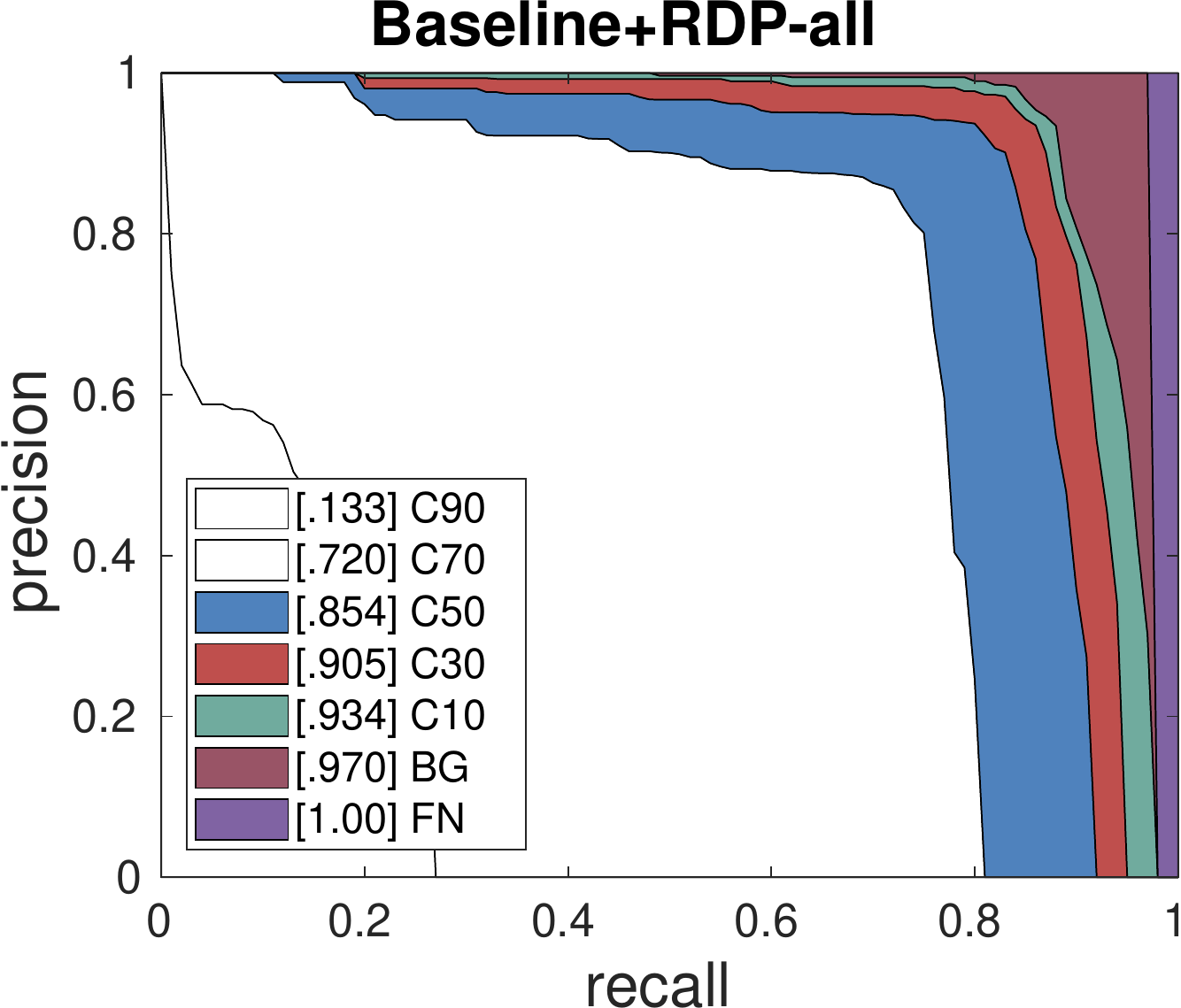} &
    \addFig{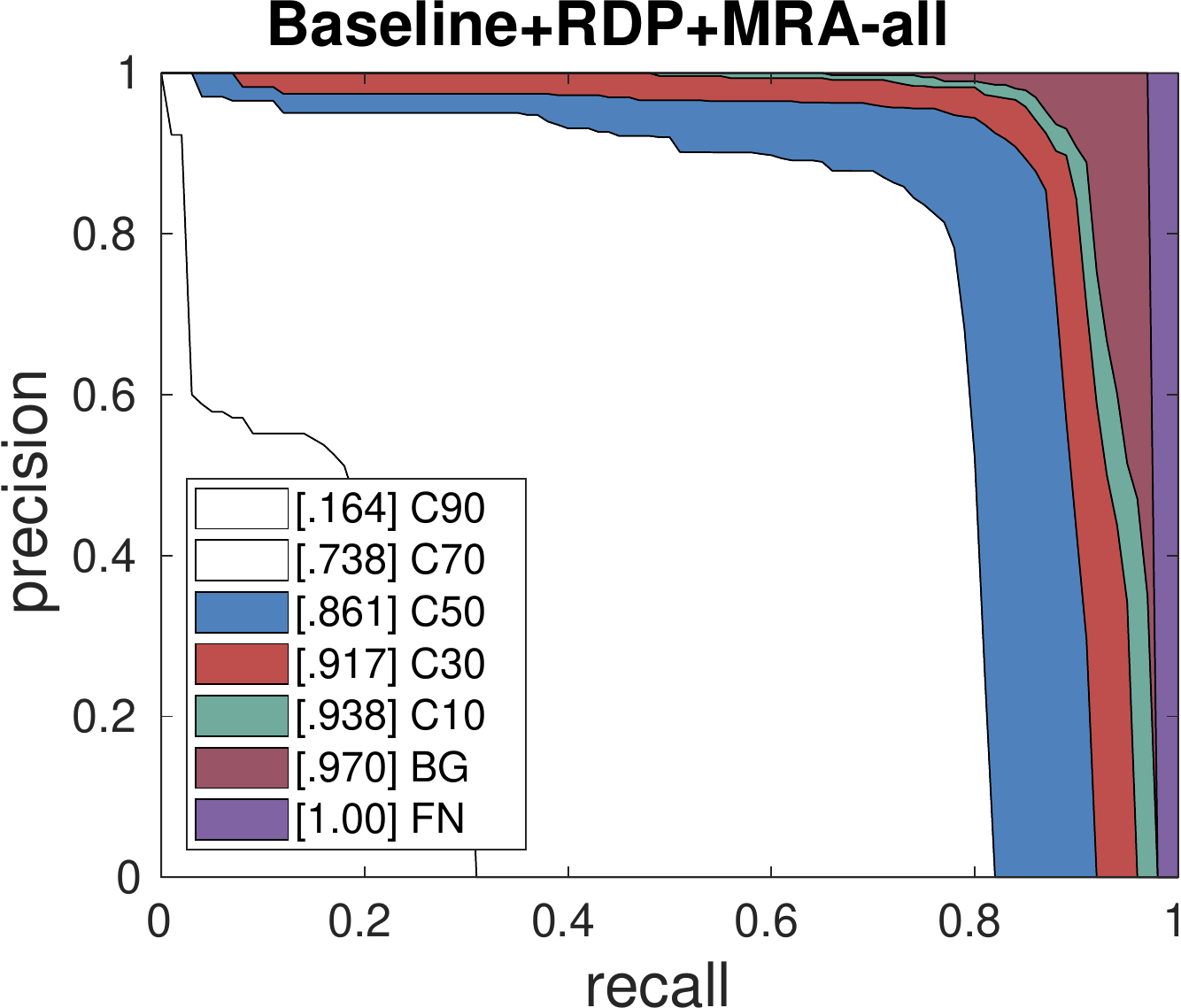} \\
    \addFig{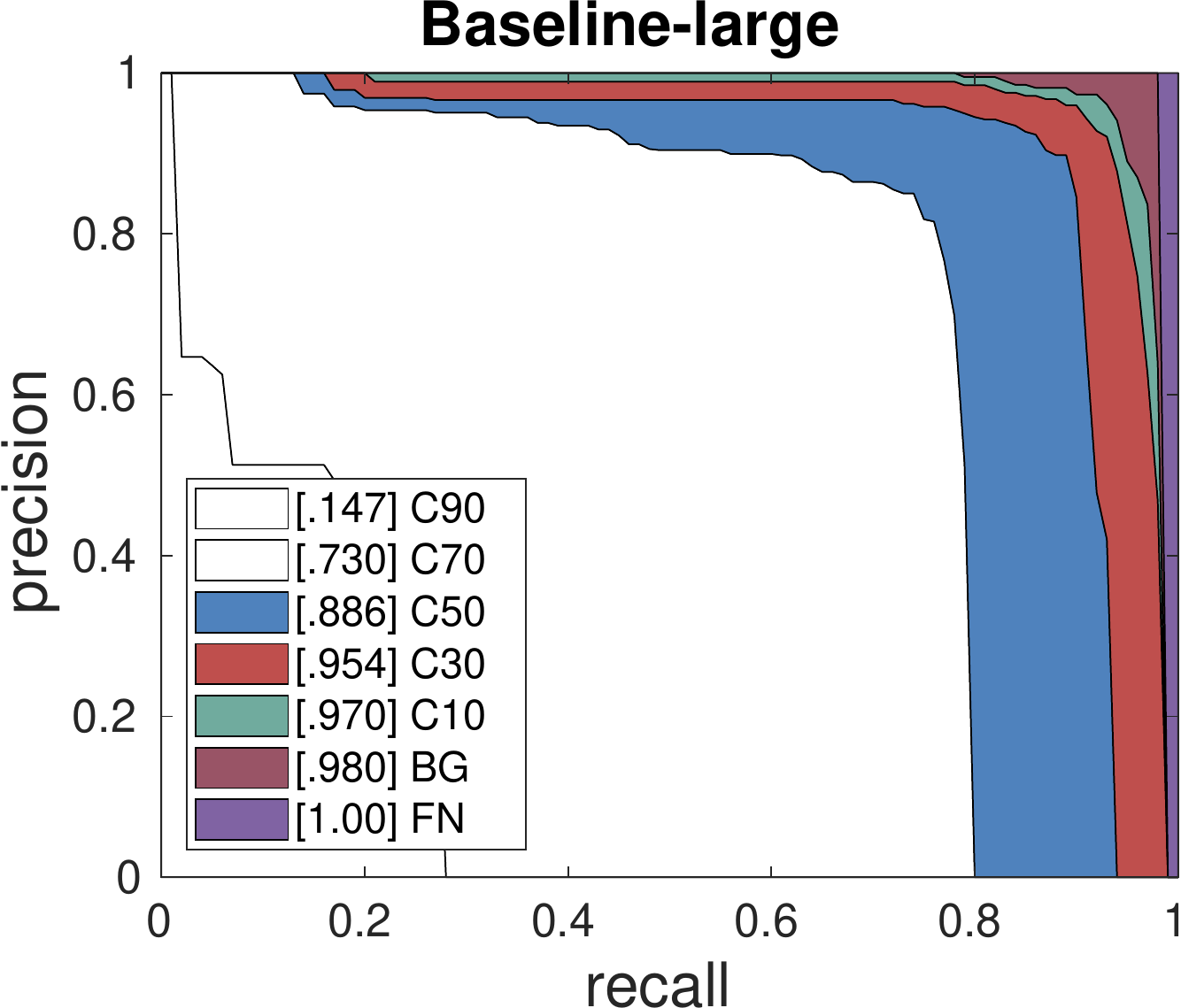} &
    \addFig{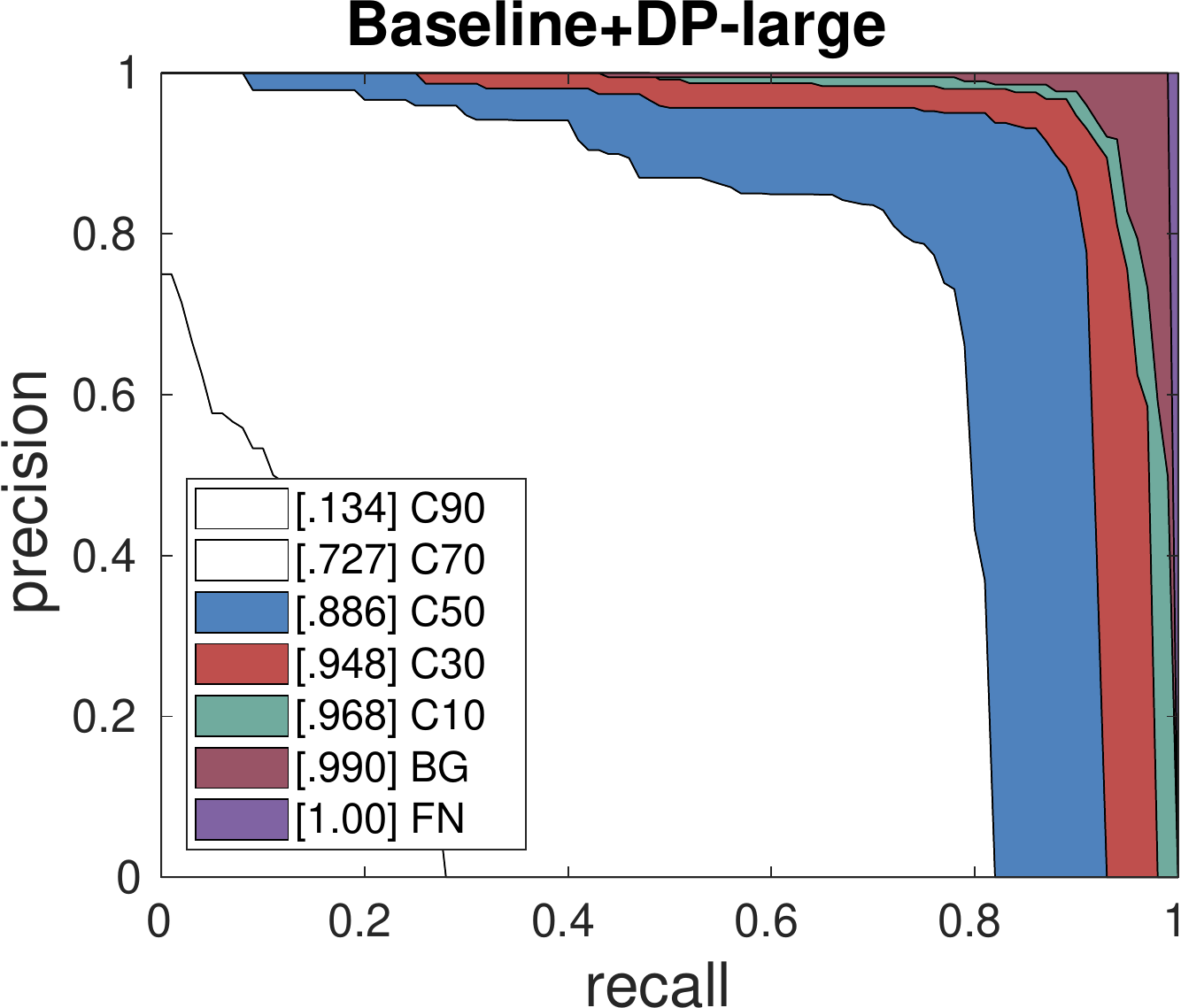} &
    \addFig{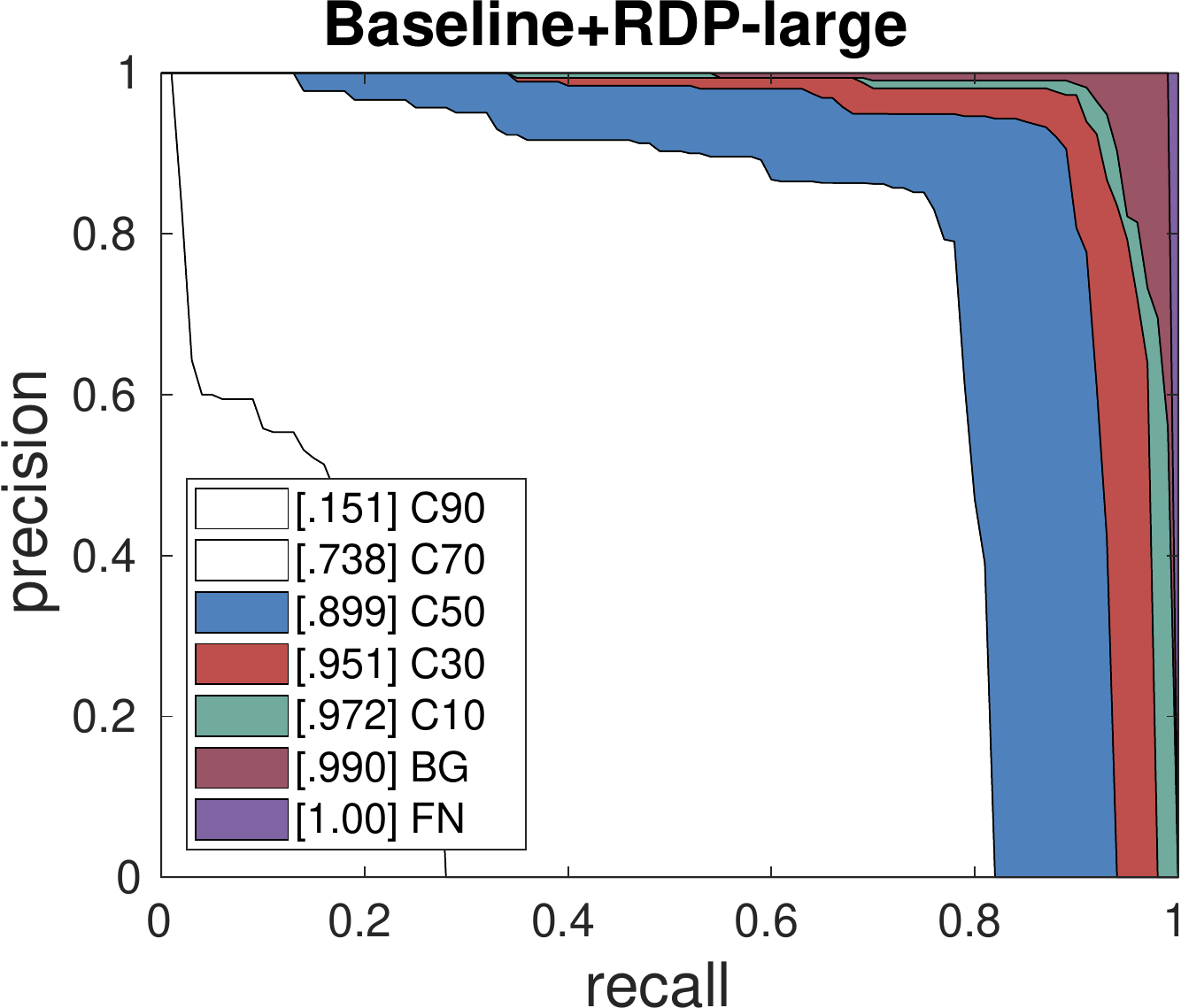} &
    \addFig{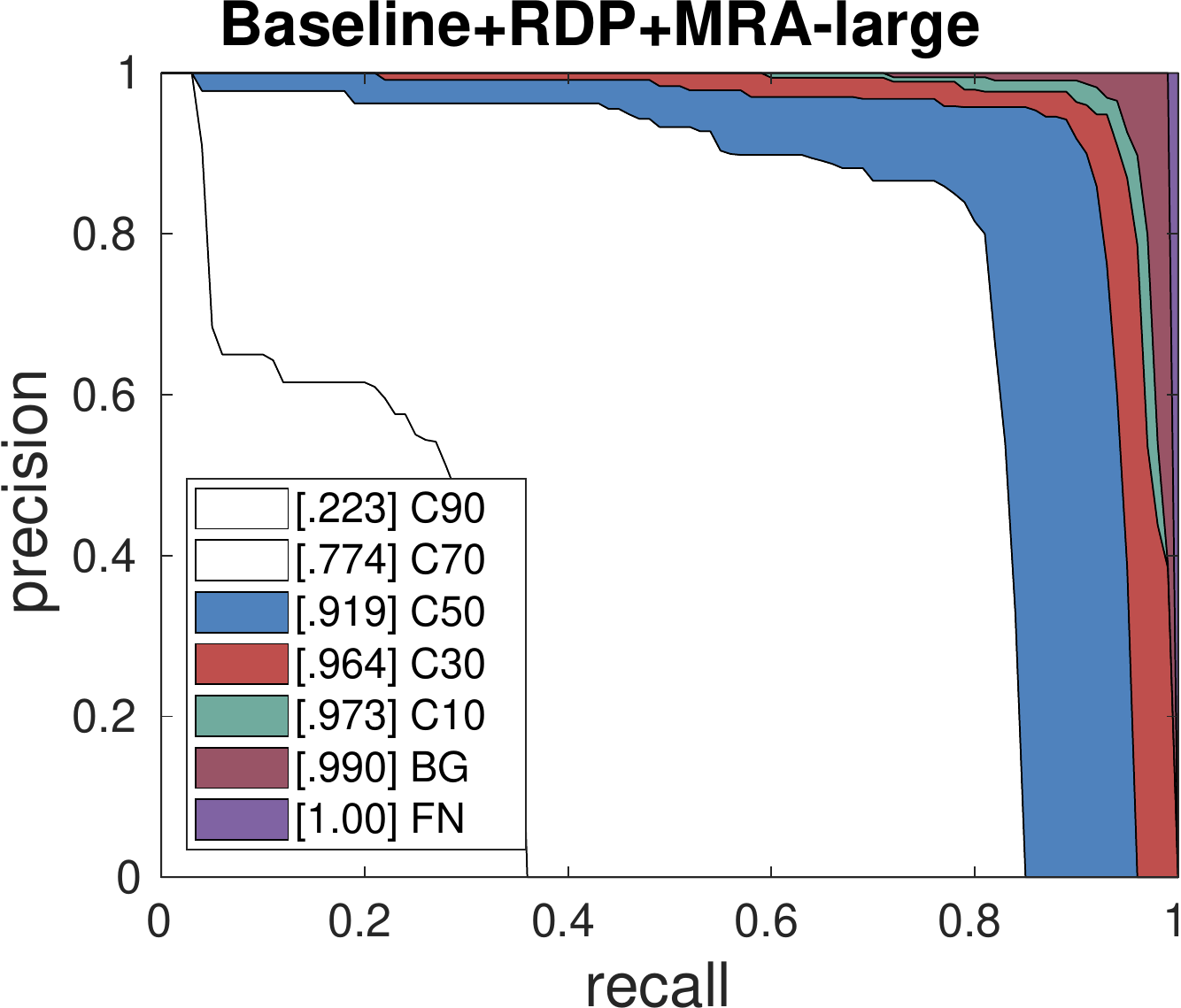} \\
\end{tabular}
\caption{Error analyses for the baseline and the proposed designs on the ISOD validation set. The first row is PR curves for all salient instances, while the second row is only for large salient instances whose areas are larger than $64^2$. The PR curves are drawn in different settings following \cite{lin2014microsoft}. C10$\sim$C90: PR curve at IoU=\{0.1:0.1:0.9\}. BG: PR curve after all false positives (FP) of background are removed. FN: PR curve after all remaining errors are removed ($\text{AP}=1$). Each number in the legend corresponds to the average precision for each setting. The area under each curve is drawn in different colors, corresponding to the color in the legend. Best viewed in color.
} \label{fig:error_ana}
\end{figure*}
}

\subsubsection{Error Analyses of the Baseline and the Proposed Designs}
Salient instances are usually large because large objects are more eye-attracting and are thus visually distinctive.
We follow the MS-COCO benchmark to consider the instances whose areas are larger than $64^2$ as large instances.
In this way, we find that the ISOD dataset~\cite{li2017instance} has over 70\% large salient instances.
Here, we perform error analyses using all salient instances or only large instances.
We view FPN \cite{lin2017feature} as the baseline and gradually add each design of us to this baseline to analyze the changes of detection errors.
\figref{fig:error_ana} illustrates the results.
First, let us discuss the changes of the PR curve by adding DP to the baseline.
We observe that although AP is improved for almost all IoU thresholds when using all salient instances, the performance becomes worse when only salient instances are considered, especially for large IoU thresholds (\eg, $\text{IoU}=0.9$).
Then, we further replace DP with the regularized version of RDP.
There is a significant improvement in terms of all IoU thresholds for both all and only large salient instances, demonstrating the importance of the proposed regularization for DP.
At last, we analyze the effect of the multi-level \textit{RoIAlign} (MRA) by further adding it to our system.
A substantial improvement is observed, especially for large salient instances.
For example, MRA brings AP improvements of 7.2\%, 3.6\%, and 2.0\% for IoU thresholds 0.9, 0.7, and 0.5, respectively.
Compared our final system (the rightmost column in \figref{fig:error_ana}) with the baseline (the leftmost column), 
the improvement is very visually significant in the PR curves in terms of all IoU thresholds.

\begin{table}[htbp]
    \centering
    \caption{\revise{Evaluation results on the ISOD \cite{li2017instance}
    and SOC \cite{fan2018salient} datasets. All methods are based on ResNet-50 except the VGG-16 based MSRNet \cite{li2017instance}}.}
    \label{tab:evaluation_results}
    \setlength{\tabcolsep}{1mm}
    \resizebox{\columnwidth}{!}{
      \begin{tabular}{l|ccc|ccc}
        \Xhline{1pt}
      \multicolumn{1}{c|}{\multirow{2}[0]{*}{Method}} & \multicolumn{3}{c|}{ISOD \cite{li2017instance}} & \multicolumn{3}{c}{SOC \cite{fan2018salient}} \\
      \cline{2-7}
            & AP & AP$_{50}$ & AP$_{70}$ & AP & AP$_{50}$ & AP$_{70}$ \\
      \hline
      MSRNet$_{17}$ \cite{li2017instance} & - & 65.3\% & 52.3\% & - & - & - \\
      MS R-CNN$_{19}$ \cite{huang2019mask} & 56.2\% & 84.2\% & 68.8\% & 35.8\% & 55.1\% & 44.2\% \\
      HTC$_{19}$ \cite{chen2019hybrid} & 45.4\% & 81.5\% & 55.9\% & 32.7\% & 57.6\% & 41.2\% \\
      CenterMask$_{20}$ \cite{lee2020centermask} & 54.0\% & 87.2\% & 68.7\% & 23.8\% & 39.5\% & 29.9\% \\
      BlendMask$_{20}$ \cite{chen2020blendmask} & 53.6\% & 88.0\% & 67.4\% & 32.3\% & 56.2\% & 38.7\% \\
      DetectoRS$_{20}$ \cite{qiao2020detectors} & 50.4\% & 82.7\% & 63.7\% & 24.3\% & 49.1\% & 28.4\% \\
      SOLO$_{20}$ \cite{wang2020solo} & 53.5\% & 84.2\% & 65.3\% & 36.0\% & 58.1\% & 45.0\% \\
      S4Net$_{20}$ \cite{fan2019s4net} & 52.3\% & 86.7\% & 63.6\% & 24.0\% & 51.8\% & 27.5\% \\
      Ours  & \textbf{58.6\%} & \textbf{88.9\%} & \textbf{73.8\%} & \textbf{37.7\%} & \textbf{59.4\%} & \textbf{48.4\%} \\
      \Xhline{1pt}
      \end{tabular}%
    }
    \label{tab:addlabel}%
  \end{table}%


\newcommand{\addFigs}[1]{!th}
\CheckRmv{
\begin{figure*}[t]
    \centering
    \small
    \renewcommand{\arraystretch}{0.5}
    \setlength{\tabcolsep}{0.25mm}
    \renewcommand{\addFig}[1]{{\includegraphics[height=.119\textwidth]{msr_examples/#1}}}
    \begin{tabular}{ccccccc}
        \rotatebox[origin=l]{90}{~~~~ Image} &
        \addFig{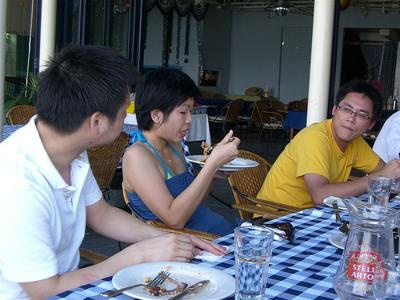} & \addFig{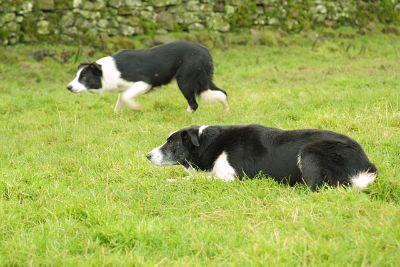} &
        \addFig{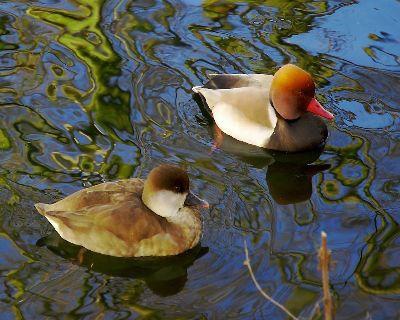} & \addFig{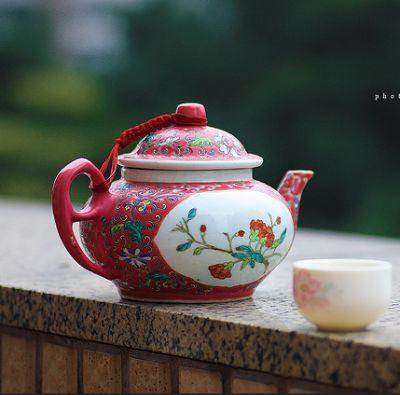} &
        \addFig{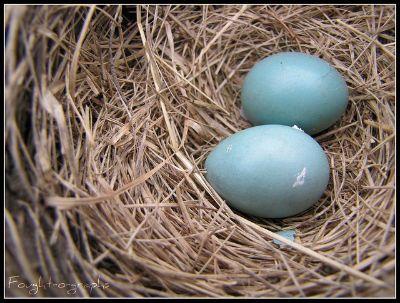} & \addFig{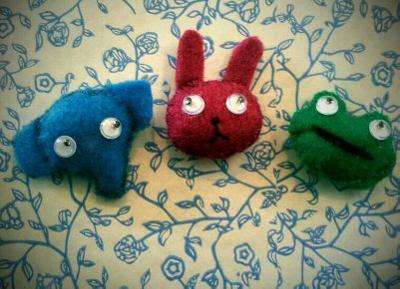}
        \\
        \rotatebox[origin=l]{90}{~ Instance GT} &
        \addFig{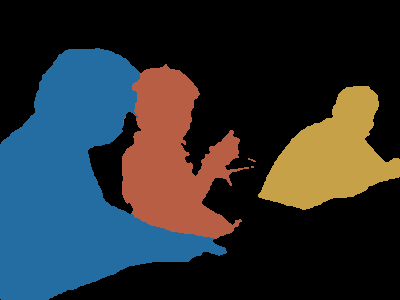} & \addFig{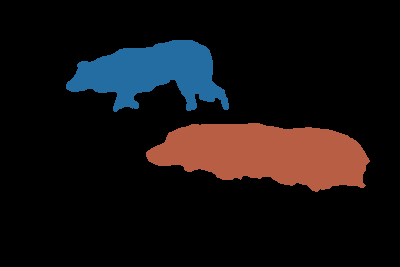} &
        \addFig{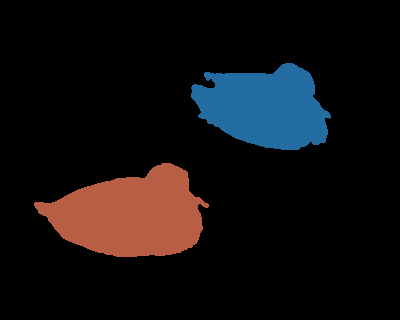} & \addFig{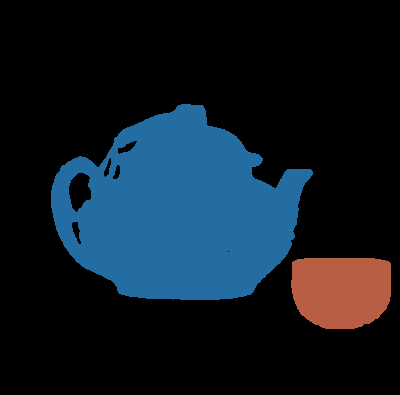} &
        \addFig{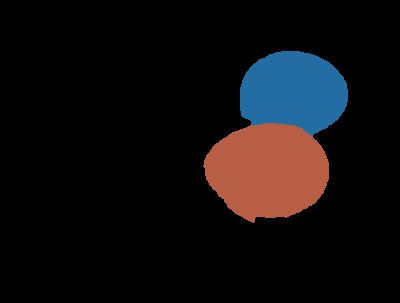} & \addFig{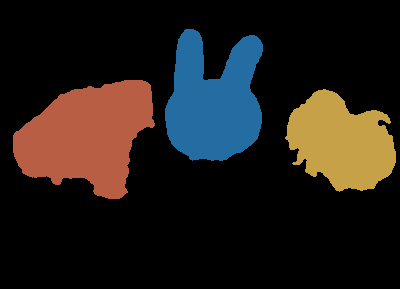}
        \\
        \rotatebox[origin=l]{90}{~~~ S4Net} &
        \addFig{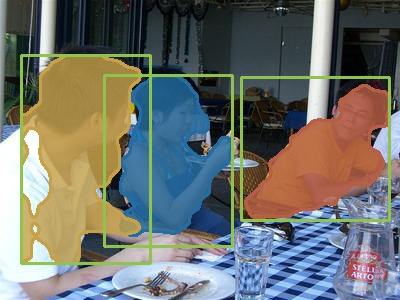} & \addFig{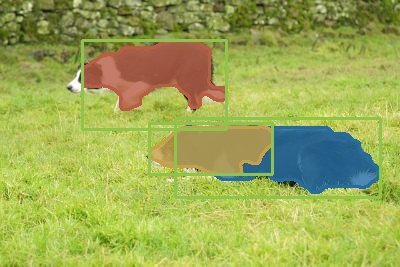} &
        \addFig{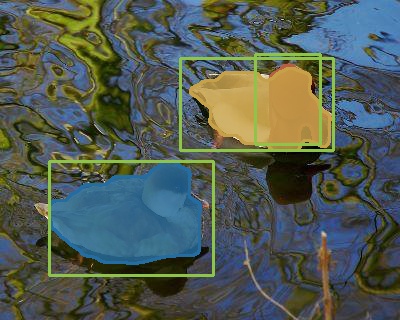} & \addFig{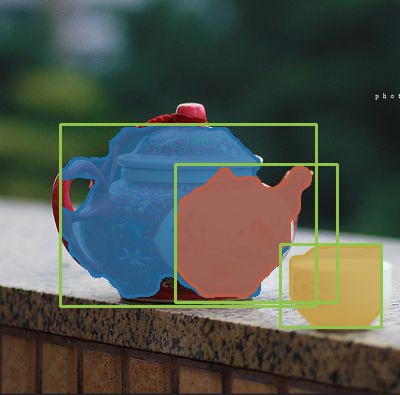} &
        \addFig{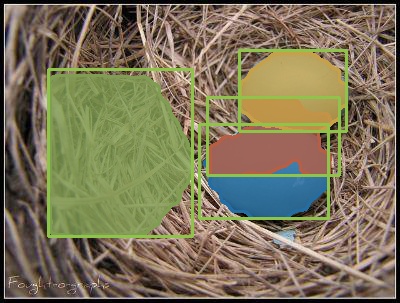} & \addFig{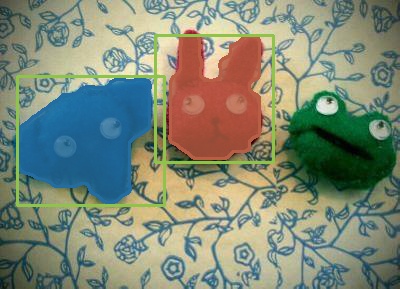}
        \\
        \rotatebox[origin=l]{90}{~~~~~ Ours} &
        \addFig{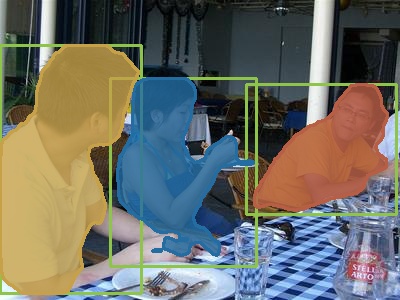} & \addFig{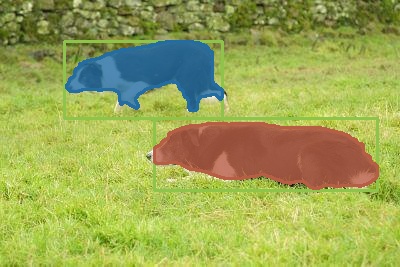} &
        \addFig{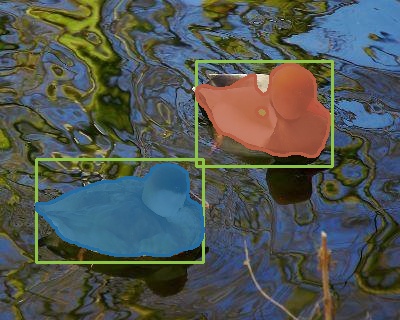} & \addFig{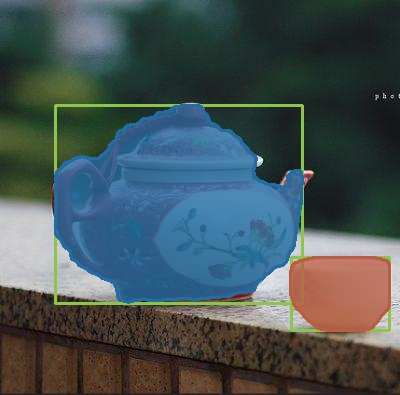} &
        \addFig{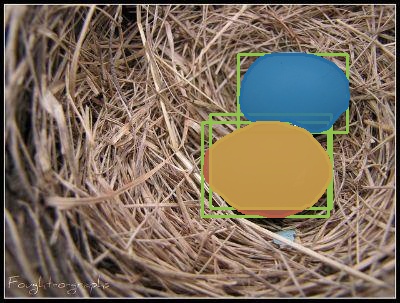} & \addFig{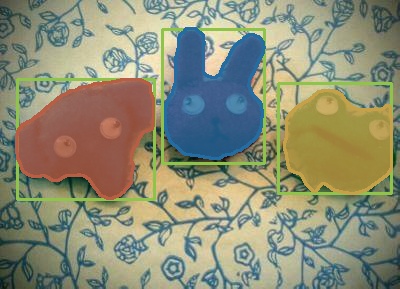}
        \\
        \multicolumn{7}{c}{\large\textbf{ISOD Dataset}}
    \end{tabular}
    \renewcommand{\addFig}[1]{{\includegraphics[height=.1163\textwidth]{soc_examples/#1}}}
    \begin{tabular}{ccccccc}
        \rotatebox[origin=l]{90}{~~~~ Image} &
        \addFig{COCO_train2014_000000071265.jpg} &
        \addFig{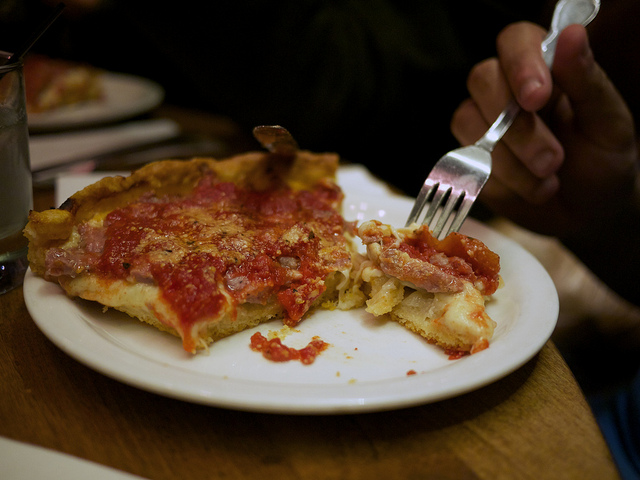} &
        \addFig{COCO_train2014_000000005115.jpg} &
        \addFig{COCO_train2014_000000006709.jpg} &
        \addFig{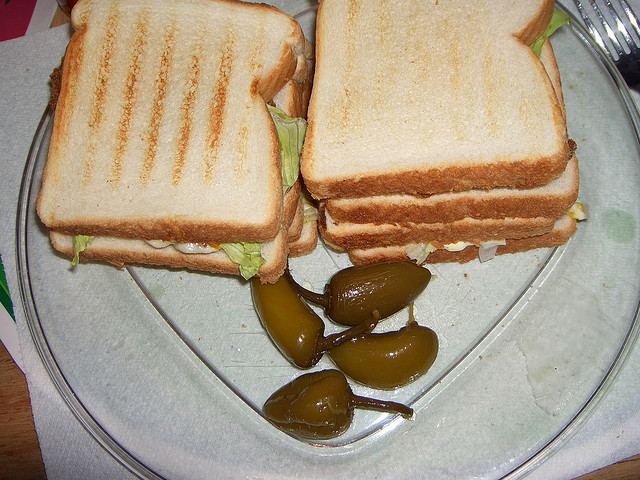} &
        \addFig{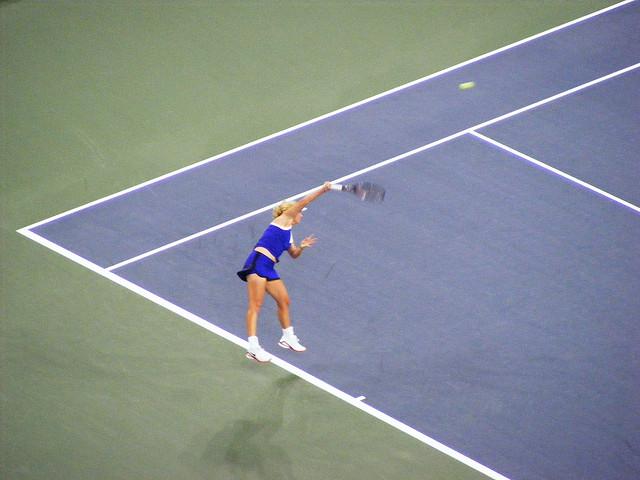}
        \\
        \rotatebox[origin=l]{90}{~ Instance GT} &
        \addFig{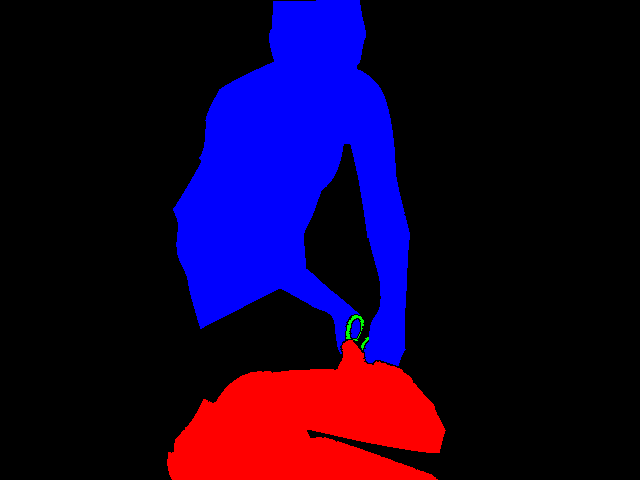} &
        \addFig{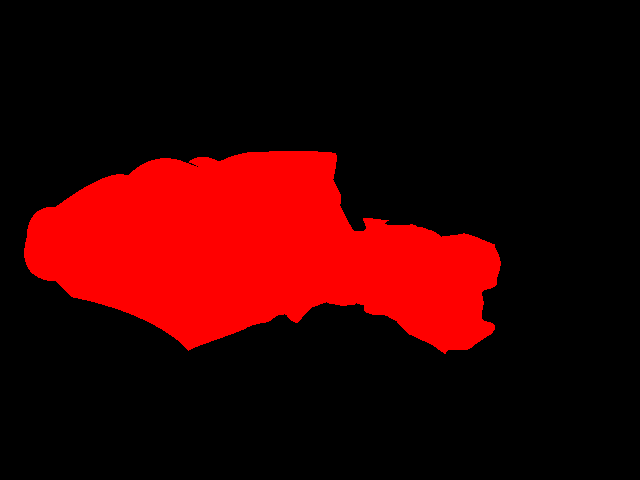} &
        \addFig{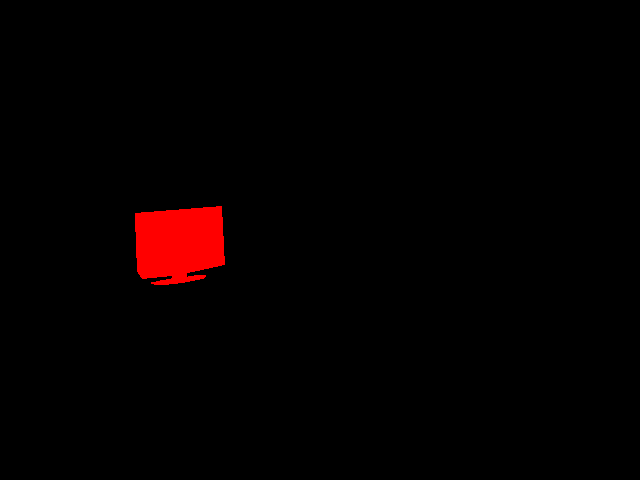} &
        \addFig{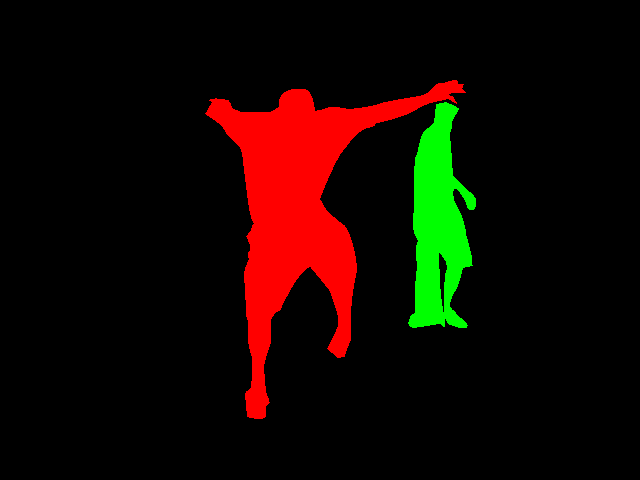} &
        \addFig{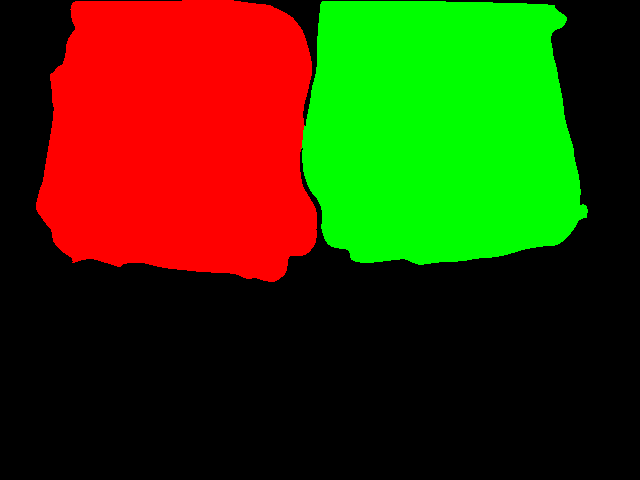} &
        \addFig{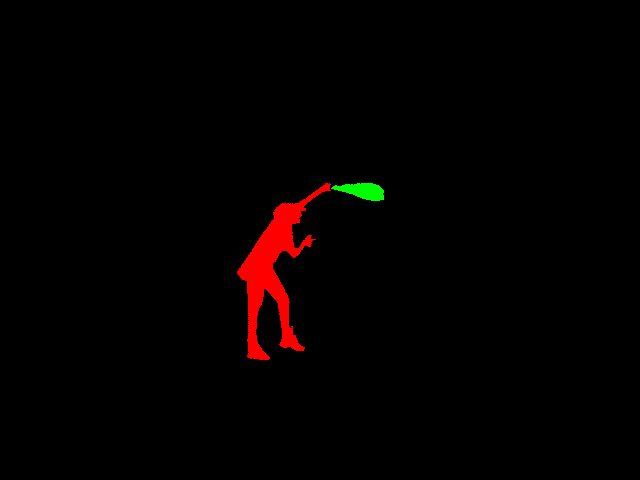}
        \\
        \rotatebox[origin=l]{90}{~~~ S4Net} &
        \addFig{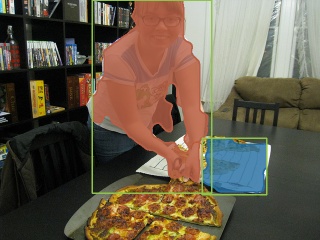} &
        \addFig{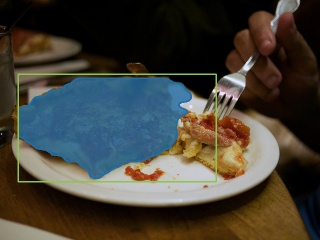} &
        \addFig{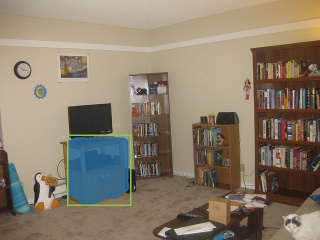} &
        \addFig{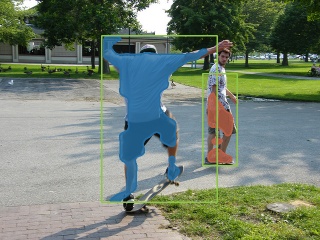} &
        \addFig{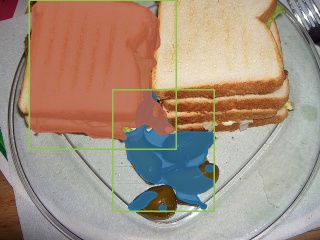} &
        \addFig{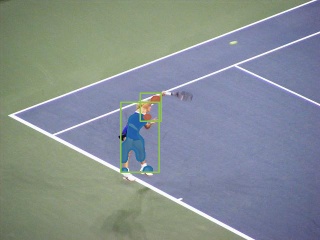}
        \\
        \rotatebox[origin=l]{90}{~~~~~ Ours} &
        \addFig{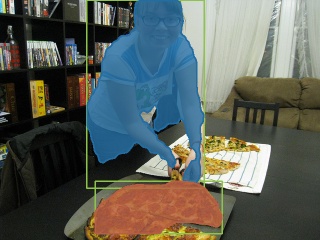} &
        \addFig{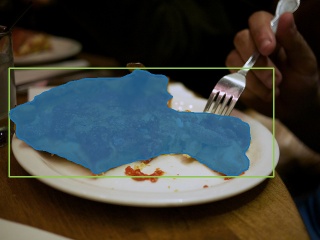} &
        \addFig{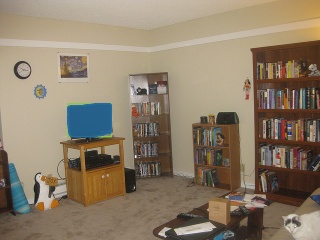} &
        \addFig{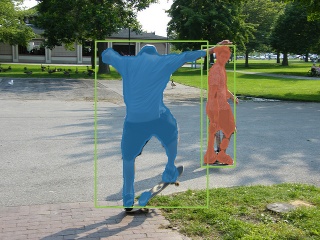} &
        \addFig{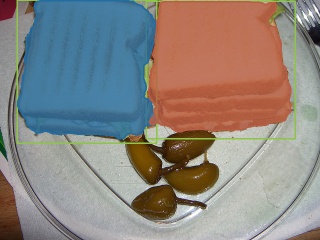} &
        \addFig{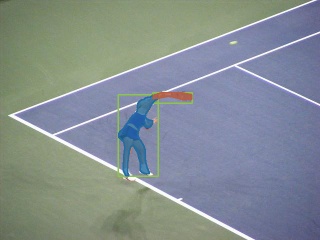}
        \\
        \multicolumn{7}{c}{\large\textbf{SOC Dataset}}
    \end{tabular}
    \caption{Qualitative comparisons between our method and S4Net~\cite{fan2019s4net}. The samples are from the ISOD and SOC datasets. S4Net~\cite{fan2019s4net} is easy to detect superfluous objects (false positives) or a part of instances. In contrast, our proposed method can detect the complete instances and have much fewer false positives.
} \label{fig:msr_examples}
\end{figure*}
}

\subsubsection{Bottom-up versus Top-down}
\label{sec:failure_case}
In our method, we rebuild the feature pyramid based on the outputs of FPN.
Another potential solution is to directly replace FPN with the top-down style of RDP, which would have a lower computational cost compared with our proposed design.
However, the experimental results proclaim its failure.
As shown in \tabref{tab:ablation_fail}, this solution leads to substantial performance degradation, \ie, over 10\% lower than the default bottom-up design in terms of various metrics.
Hence, 
we can come to the conclusion that the proposed RDP is not suitable for the top-down information flow but can only well in a bottom-up way.

\subsubsection{\revise{Feature Pyramid Enhancement Strategies}}
\revise{
\tabref{tab:fpn_cmp} shows the quantitative comparison of the proposed RDP with other competitive feature pyramid enhancement strategies, \ie, PA \cite{liu2018path}, NAS-FPN \cite{ghiasi2019fpn}, and BiFPN \cite{tan2020efficientdet}.
We use the same baseline as in \secref{sec:failure_case}.
One can see that PA \cite{liu2018path}, NAS-FPN \cite{ghiasi2019fpn}, and BiFPN \cite{tan2020efficientdet} have a minor improvement (0.4\% for PA \cite{liu2018path}, 0.1\% for NAS-FPN \cite{ghiasi2019fpn})
or even no improvement ($-$0.1\% for BiFPN \cite{tan2020efficientdet}) over the baseline, in terms of the AP metric.
In contrast, our proposed RDP outperforms the baseline by a large margin (2.2\% AP improvement), demonstrating its superiority in feature pyramid enhancement.
}

\CheckRmv{
\begin{table}[!t]
    \centering
    \renewcommand{\tabcolsep}{3.0mm}
    \caption{Evaluation of our method with different backbone networks on the ISOD test set \cite{li2017instance}. 
    Our method with the most powerful backbone (\ie,
    ResNeXt-101 \cite{xie2017aggregated}) can achieve a 4.6\% improvement in terms of AP and 2.7$\times$ inference time 
    compared with that with the simplest backbone (\ie, ResNet-50 \cite{he2016deep}). 
    The speed is tested using a single NVIDIA TITAN Xp GPU.
    } \label{tab:eval_backbone}
    \begin{tabular}{c|cccc} \hline
        Backbone & AP & AP$_{50}$ & AP$_{70}$ & Speed \\ \hline\hline
        ResNet-50 \cite{he2016deep} & 58.6\% & 88.9\% & 73.8\% & 45.0fps \\
        ResNet-101 \cite{he2016deep} & 60.9\% & 89.7\% & 76.6\% & 34.8fps \\
        ResNeXt-101 \cite{xie2017aggregated} & 63.2\% & 90.1\% & \textbf{78.1\%} & 16.7fps \\ 
        \hline
      \end{tabular}
\end{table}
}

\subsection{Comparisons with \sArt Methods} \label{comparison}
Since SIS is a relatively new problem, the previous works on this topic are very limited.
Here, we compare our method with two well-known SIS methods: MSRNet~\cite{li2017instance} that is on behalf of the post-processing-based methods and S4Net~\cite{fan2019s4net} that is a representative work of end-to-end networks.
\revise{Moreover, we compare our method with recent well-known instance segmentation methods,
including
Mask Scoring (MS) R-CNN \cite{huang2019mask}, HTC \cite{chen2019hybrid},
CenterMask \cite{lee2020centermask},
BlendMask \cite{chen2020blendmask},
SOLO \cite{wang2020solo}, and DetectoRS \cite{qiao2020detectors}.
For a fair comparison, we train all the above methods using their official code with default settings and the ResNet-50 backbone.}
Hence, all methods are based on ResNet-50 \cite{he2016deep} except the VGG-16 based MSRNet \cite{li2017instance}.

\subsubsection{ISOD Dataset}
Following \cite{li2017instance,fan2019s4net}, all methods are tested on the ISOD test set~\cite{li2017instance}.
The quantitative results can be seen in \tabref{tab:evaluation_results}.
\revise{The proposed method achieves the best results compared with the other two popular competitors and recent strong instance segmentation methods.}
Specifically, the proposed method has 6.3\% higher AP than S4Net~\cite{fan2019s4net}.
In terms of AP$_{70}$, the proposed method is 10.2\% better than S4Net~\cite{fan2019s4net}.
\revise{Compared with recent strong instance segmentation methods,
our method has a significant 2.4\% improvement in terms of the AP metric.
These results demonstrate the superiority of the proposed method in accurate SIS.}
In \tabref{tab:eval_backbone}, we try different backbone networks for our method.
One can see that powerful backbones can further significantly boost the performance,
indicating the good potential and extendibility of our method.

\subsubsection{SOC Dataset}
The SOC dataset scenarios~\cite{fan2018salient} are much more complex than those of the ISOD dataset~\cite{li2017instance}, so SIS on the SOC dataset is more challenging.
The quantitative comparison between our method and other recent methods
on the SOC dataset is summarized in \tabref{tab:evaluation_results}.
Since other methods do not report evaluation results on this dataset, we train S4Net~\cite{fan2019s4net} using its official code with default settings.
\revise{We leave the performance of MSRNet \cite{li2017instance} blank 
due to its incomplete code.}
The results suggest that our method is 13.7\%, 7.6\%, and 20.9\% better than S4Net in terms of AP, AP$_{50}$, and AP$_{70}$, respectively.
\revise{Compared with recent strong instance segmentation methods,
our method still has 1.7\%, 1.3\%, 3.4\% improvement in terms of AP, AP$_{50}$, and AP$_{70}$, respectively.}
The above result suggests that our method can handle the cluttered background much better and our improvement for SIS is nontrivial.

\renewcommand{\addFig}[1]{{\includegraphics[width=.24\textwidth]{curve/sta_ana/#1.pdf}}}

\CheckRmv{
\begin{figure}[t]
    \centering
    \renewcommand{\arraystretch}{1.0}
    \setlength{\tabcolsep}{0.2mm}
    \begin{tabular}{cc}
        \addFig{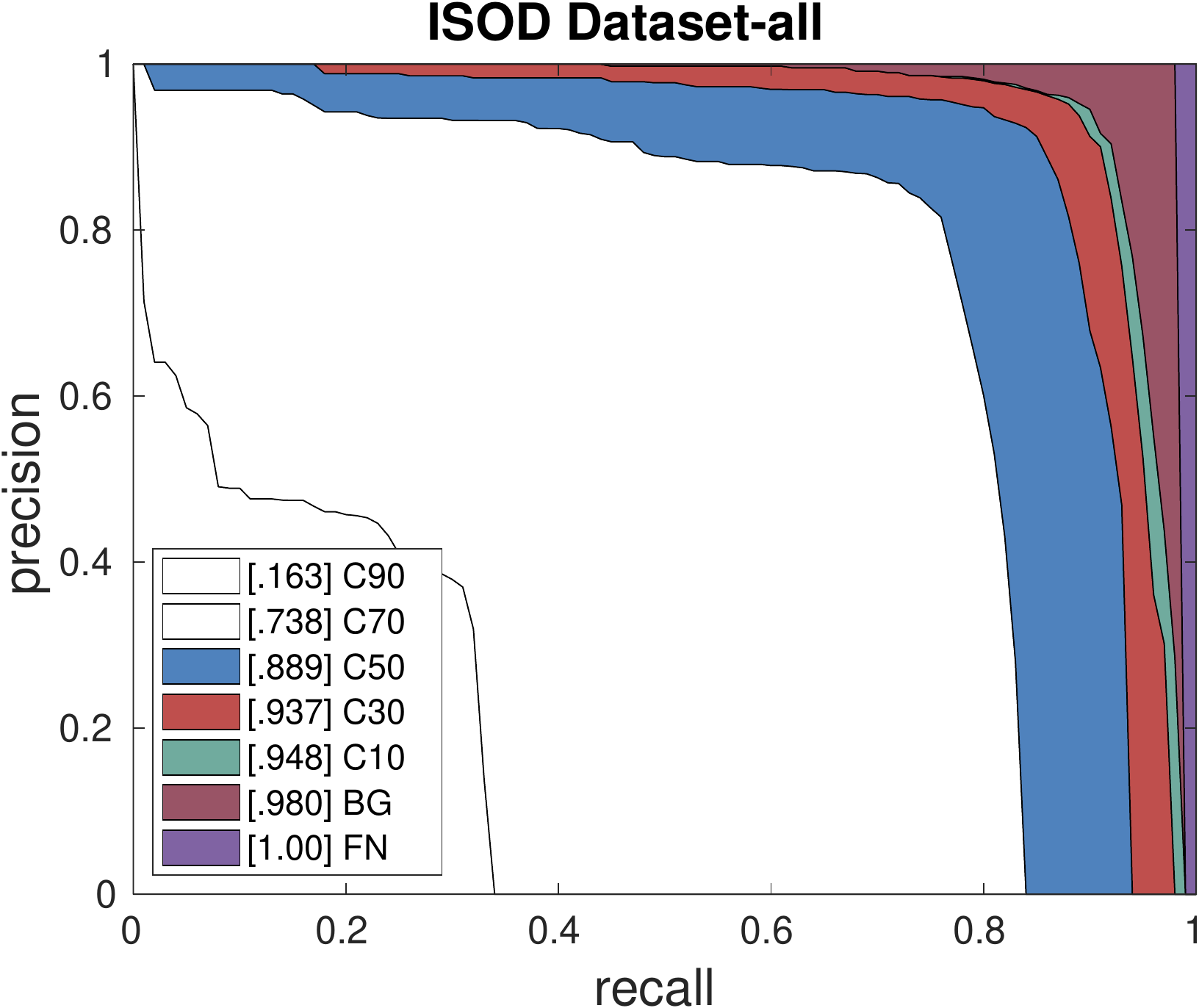} & \addFig{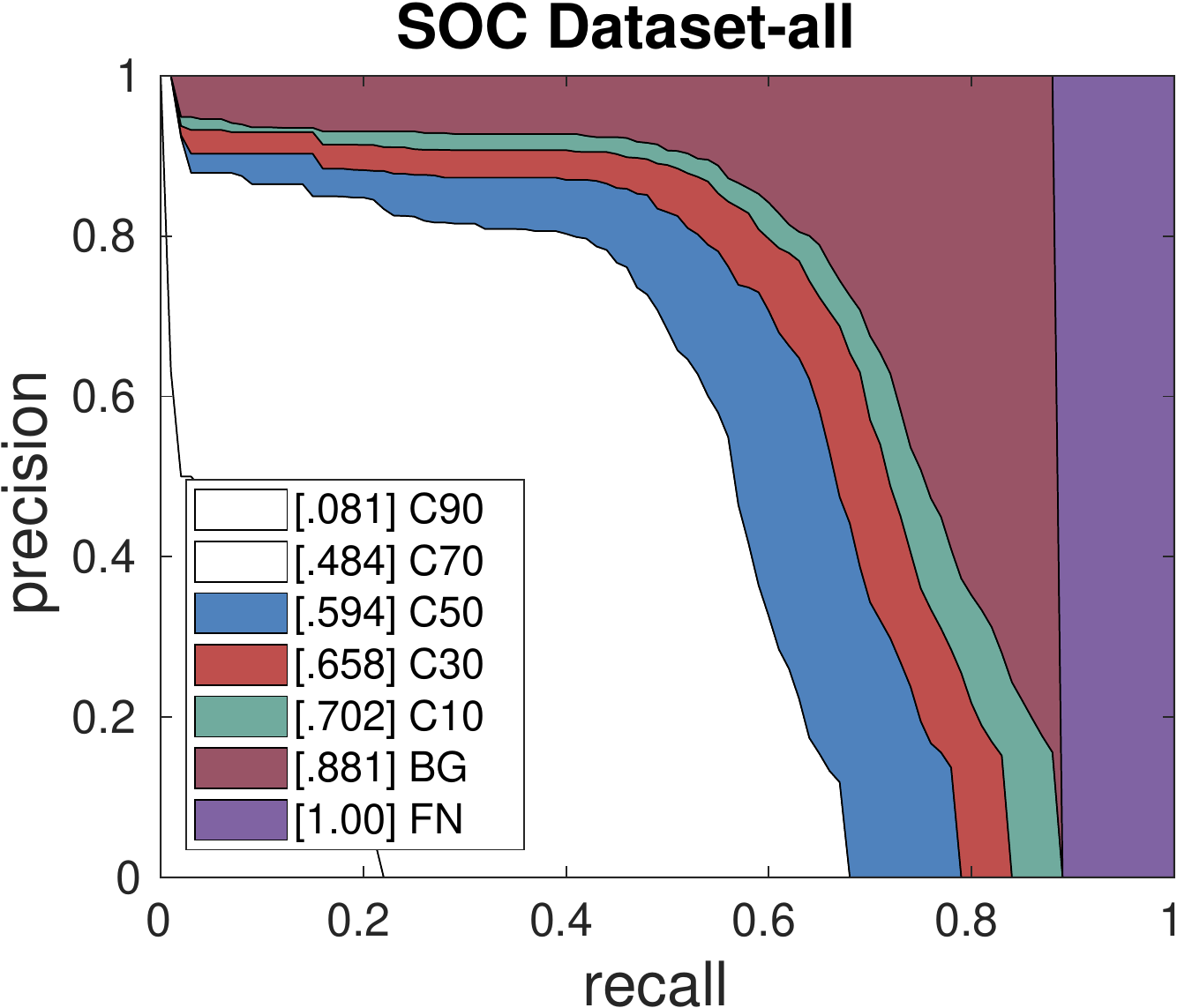} \\
        \multicolumn{2}{c}{\footnotesize (a) Error analyses} \\
    \end{tabular}
    \begin{tabular}{cc}
        \addFig{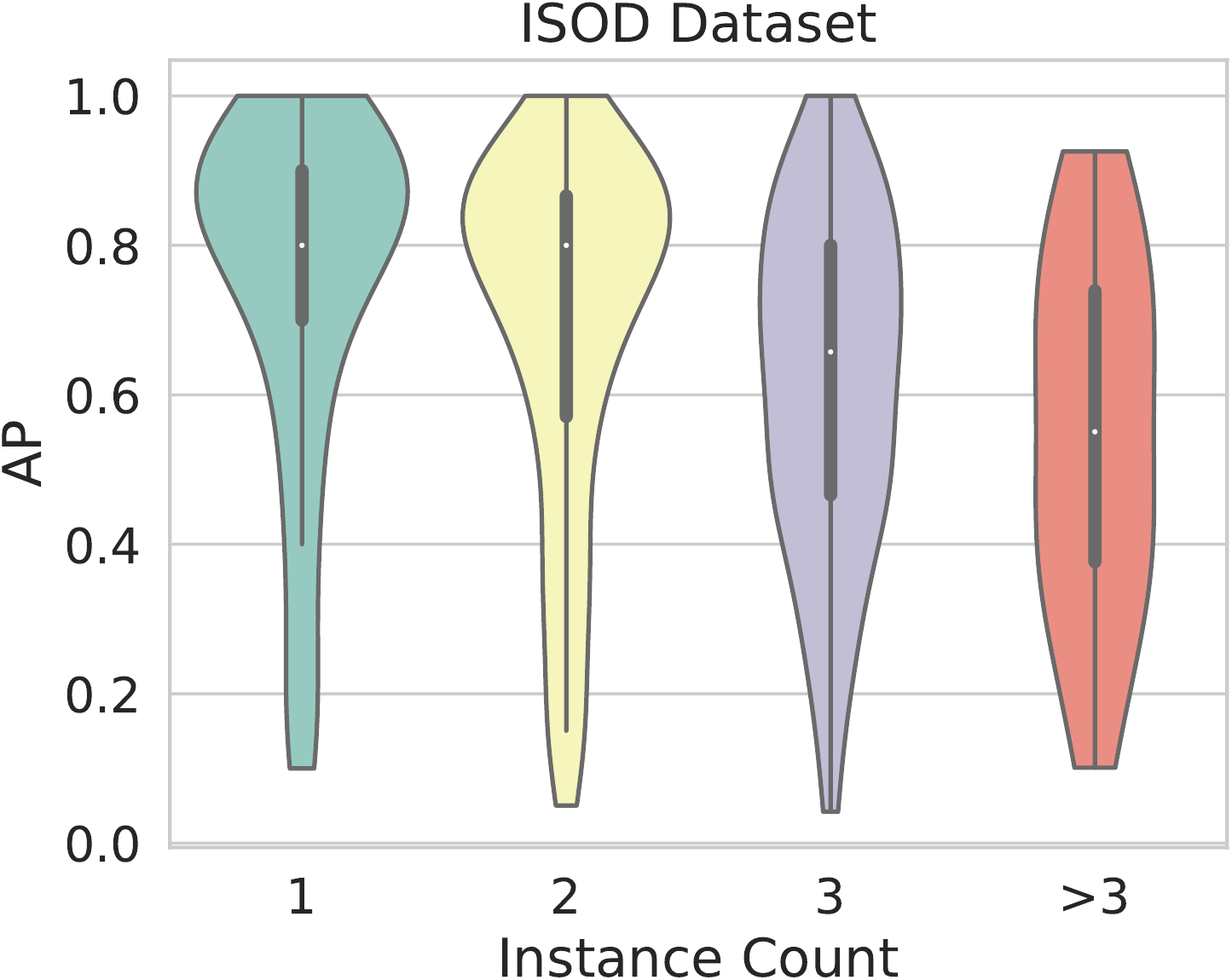} & \addFig{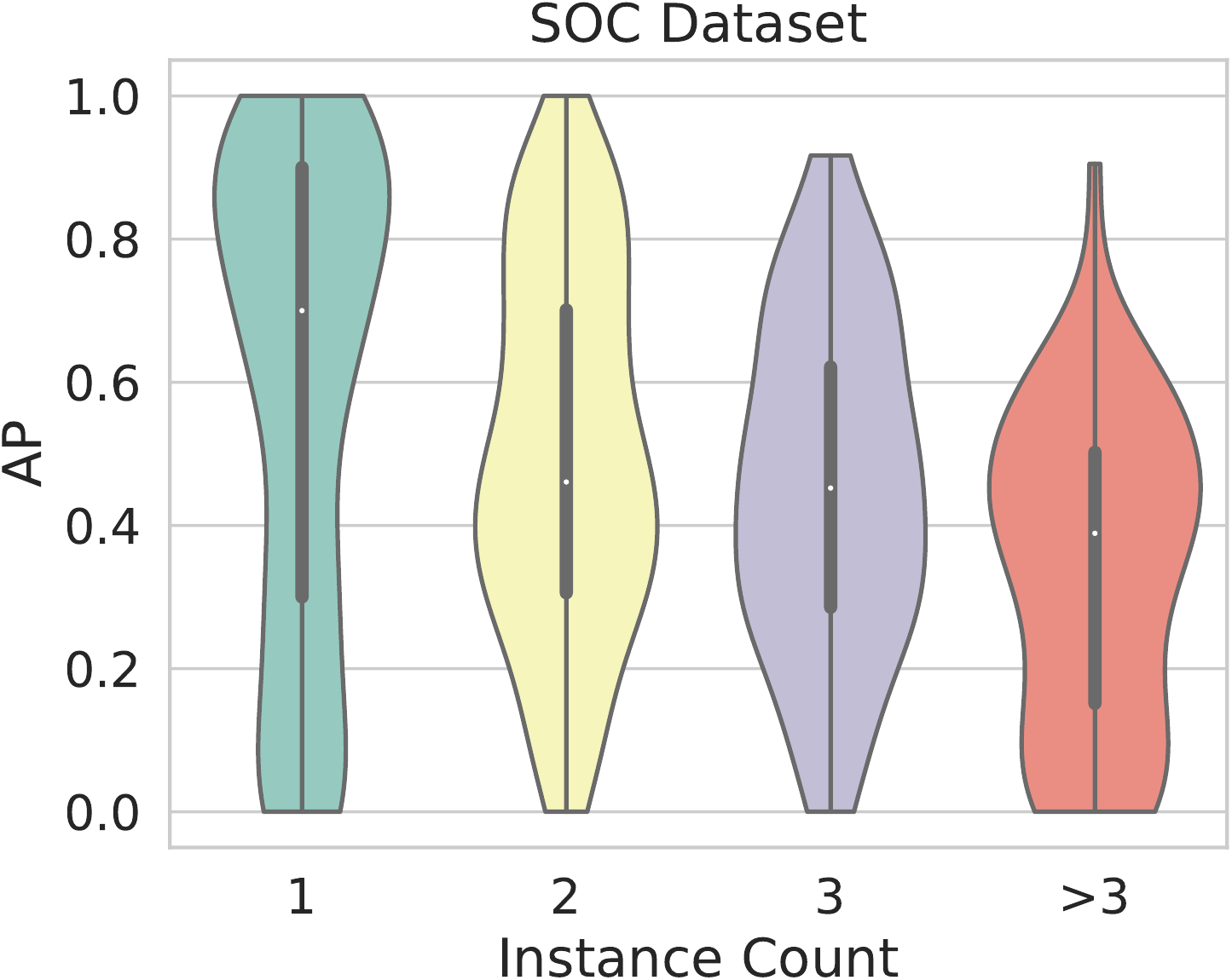} \\
        \multicolumn{2}{c}{\footnotesize (b) The probability distribution of AP} \\
    \end{tabular}
    \caption{Statistical analyses for our method on the ISOD \cite{li2017instance} and SOC \cite{fan2018salient} test sets.}
    \label{fig:sta_ana}
\end{figure}
}

\subsection{Qualitative Comparisons}
To visually compare our method with the previous \sArt method of S4Net \cite{fan2019s4net}, we show qualitative comparisons using the ISOD \cite{li2017instance} and SOC \cite{fan2018salient} datasets in \figref{fig:msr_examples}.
S4Net has many superfluous detection results (false positives) or only detects a part of salient instances.
In contrast, our method produces consistent and high-quality results.
Moreover, the boundaries of salient instances detected by S4Net are usually rough, while our method can produce salient instances with smooth boundaries.
Therefore, these qualitative comparisons further validate the effectiveness of the proposed method.

\subsection{Statistical Analyses}
The statistical characteristics of the ISOD \cite{li2017instance} and SOC \cite{fan2018salient} datasets are highly different, so it would be interesting to explore the differences in the performance of our method on these two datasets.
Here, we conduct statistical analyses for the performance of our method on the test sets of these two datasets.
We first explore the differences of PR curves between the two datasets by drawing the PR curves of our method on these two datasets, as shown in \figref{fig:sta_ana} (a).
As the background of images in the SOC dataset is more cluttered than that in the ISOD dataset, more salient instances are not detected in the SOC dataset.
In contrast, in the ISOD dataset, most salient instances can be correctly localized.
Then, we explore the probability distribution of AP for different numbers of salient instances in each image.
More specifically, we calculate the AP score and the number of ground-truth salient instances for each image. 
We then illustrate the overall probability distribution in \figref{fig:sta_ana} (b), where the area of each closed pattern is 1 (\ie, the sum of all probabilities).
$\text{AP}=1$ for an image means that our method almost perfectly detects and segments the ground truths in this image and has no false positives.
$\text{AP}=0$ indicates that all ground truths in this image are not detected.
In the ISOD dataset, each image's AP score is likely better than the medium AP score if the instance count is not more than 3 in each image,
while in the SOC dataset, the same case happens only when the instance count is 1 in each image.
Besides, in the ISOD dataset, our method only fails for a few images ($\text{AP}=0$) with 1 or 2 salient instances in each image. 
However, in the SOC dataset, our method fails for relatively many more images.
The above analyses suggest that the SOC dataset is much more difficult than the ISOD dataset due to its cluttered background and complex scenarios. 
Therefore, there might still be much space to strengthen the representation for future SIS research.

\section{Conclusion and Future Work}
In this paper, we propose a new network for salient instance segmentation (SIS).
Our method's core is the regularized dense-connected pyramid (RDP), 
which provides each side-output with richer yet more compatible bottom-up information flows to enhance the side-output prediction.
We further design a novel multi-level RoIAlign based decoder for better mask prediction.
Through extensive experiments, we analyze the effect of our proposed designs and demonstrate the effectiveness of our method.
With our simple designs, the proposed method achieves \sArt results on popular benchmarks in terms of all evaluation metrics while keeping a real-time speed.
The effectiveness and efficiency of the proposed method make it possible for many real-world applications.
Moreover, this research is expected to push forward the development of feature learning and mask prediction for SIS.
In the future, we plan to apply the RDP module for other vision tasks that need powerful feature pyramids.
To promote future research,
code and pretrained models will be released at \url{https://github.com/yuhuan-wu/RDPNet}.

\section*{Acknowledgment}

This research was supported by 
the Major Project for New Generation of AI under Grant No. 2018AAA0100400,
S\&T innovation project from Chinese Ministry of Education,
and NSFC (61922046).

{\small
\bibliographystyle{IEEEtran}
\bibliography{reference}
}

\newcommand{\AddPhoto}[1]{\vspace{-.1in} \includegraphics[width=1in]{#1}}

\vspace{-.2in}
\begin{IEEEbiography}[\vspace{-.1in} \AddPhoto{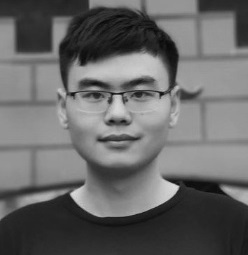}]{Yu-Huan Wu}
is currently a Ph.D. candidate with College of Computer Science
at Nankai University, supervised by Prof. Ming-Ming Cheng.
He received his bachelor's degree from Xidian University in 2018.
His research interests include computer vision
and machine learning.
\end{IEEEbiography}

\vspace{-.2in}
\begin{IEEEbiography}[\vspace{-.1in} \AddPhoto{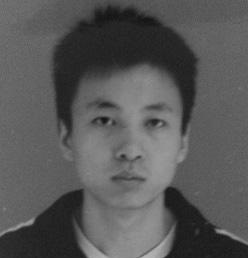}]{Yun Liu}
  received his Ph.D. degree from Nankai University in 2020.
  Currently, he works as a postdoctoral scholar with Prof. Luc Van Gool in ETH Zurich.
  His research interests include computer vision and machine learning.
  He has published over 10 papers in IEEE TPAMI, IJCV, 
  IEEE CVPR, IEEE ICCV, \etc. 
  He has received the National Scholarship in 2017, 2019, and 2020 three times.
\end{IEEEbiography}

\vspace{-.2in}
\begin{IEEEbiography}[\AddPhoto{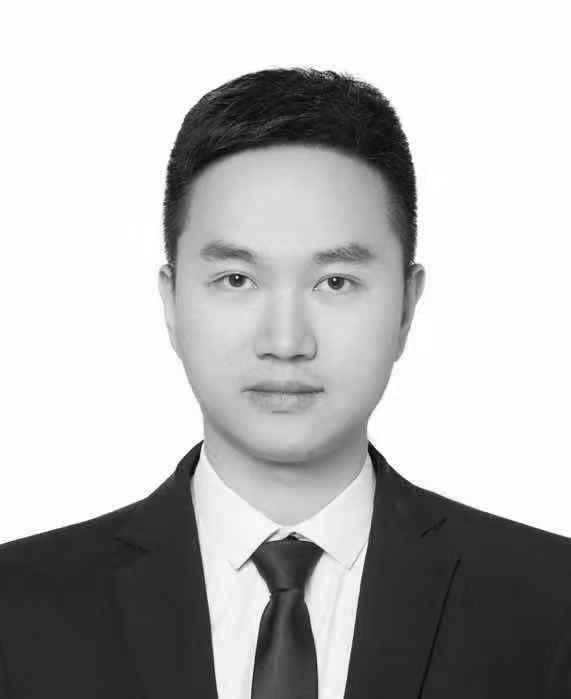}]{Le Zhang}
received his M.Sc and Ph.D.degree form Nanyang Technological University
(NTU) in 2012 and 2016, respectively.
Currently, he is a scientist at Institute for Infocomm Research,
Agency for Science, Technology and Research (A*STAR), Singapore.
He served as TPC member in several conferences such as AAAI, IJCAI.
He has served as a Guest
Editor for Pattern Recognition and Neurocomputing;
His current research interests include deep learning and computer vision.
\end{IEEEbiography}

\vspace{-.2in}
\begin{IEEEbiography}[\vspace{-.1in} \AddPhoto{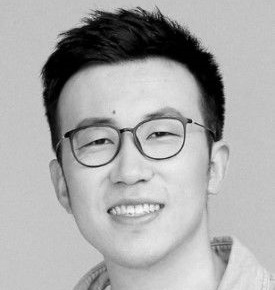}]{Wang Gao}
received his master degree from The Third Research Institute
of China Aerospace Science and Industry Corporation in 2017.
He is currently with the Science and Technology on Complex
System Control and Intelligent Agent Cooperation Laboratory,
Beijing, China.
His research interests include computer vision,
scene matching and visual navigation.
\end{IEEEbiography}

\vspace{-.2in}
\begin{IEEEbiography}[\AddPhoto{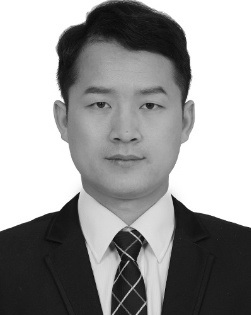}]{Ming-Ming Cheng}
received his PhD degree from Tsinghua University in 2012.
Then he did two years research fellow with Prof. Philip Torr
in Oxford.
He is now a professor at Nankai University, leading the
Media Computing Lab.
His research interests include computer graphics, computer
vision, and image processing.
He received research awards, including ACM China Rising Star Award,
IBM Global SUR Award, and CCF-Intel Young Faculty Researcher Program.
He is on the editorial boards of IEEE TIP.
\end{IEEEbiography}

\vfill

\end{document}